\documentclass[11pt]{article}

\usepackage{acl}

\usepackage{times}
\usepackage{latexsym}

  \usepackage{array}

\usepackage{amsmath}
\usepackage[dvipsnames]{xcolor}
\usepackage[framemethod=tikz]{mdframed}
\usepackage{multirow}
\usepackage{booktabs}
\usepackage{adjustbox}

\usepackage[T1]{fontenc}

\usepackage[utf8]{inputenc}
\usepackage{csquotes} 

\usepackage{microtype}

\usepackage{inconsolata}

\usepackage{graphicx}

\usepackage{amsmath}
\usepackage{amssymb}
\usepackage{amsthm}
\usepackage{graphicx}
\usepackage{booktabs}
\usepackage{enumitem}
\usepackage{algorithm}
\usepackage{algorithmic}
\usepackage{subcaption}
\usepackage{multirow}
\usepackage{tcolorbox}
\usepackage{tikz}
\usepackage{caption}
\usepackage{booktabs}
\usepackage{multirow}
\usepackage{colortbl}
\definecolor{paleblue} {RGB}{214,234,255}
\definecolor{palegrey} {RGB}{225,225,225}
\definecolor{palegreen}{RGB}{214,245,214}
\definecolor{palered}  {RGB}{255,214,214}
\usepackage{cleveref}

\usepackage{tcolorbox}
\tcbuselibrary{skins}
\usepackage{xparse}
\usepackage{etoolbox}

\newif\ifdraftmode
\draftmodetrue   

\definecolor{commentcolor}{RGB}{130, 169, 209}
\definecolor{todocolor}{RGB}{132,176,107}

\NewTColorBox{todobox}{m}{%
  enhanced,
  sharp corners,
  boxrule=0.6pt,
  colframe=todocolor,
  colback=todocolor!6,
  left=6pt,right=6pt,top=4pt,bottom=4pt,
  fontupper=\footnotesize,
}

\usepackage[dvipsnames]{xcolor}
\definecolor{aigengray}{gray}{0.45}

\newtcolorbox{commentblock}{
colback=gray!8,
colframe=gray!60,
boxrule=0.4pt,
arc=3pt,
left=6pt,
right=6pt,
top=4pt,
bottom=4pt,
fontupper=\footnotesize,
title=\textbf{Author Notes},
coltitle=black,
}

\newtheorem{definition}{Definition}

\title{
Is Memorization Context-Sensitive? Prefix-Based Extraction Beyond Isolated Prefixes
}

\author{
Ali Satvaty\textsuperscript{1} \quad 
Narjes Sharafi\textsuperscript{1} \quad 
Jirui Qi\textsuperscript{1} \quad 
Suzan Verberne\textsuperscript{2} \quad 
Fatih Turkmen\textsuperscript{1} \\
\textsuperscript{1}University of Groningen \quad
\textsuperscript{2}Leiden University \\
\textit{Correspondence:} \texttt{a.satvaty@rug.nl}
}

\begin{document}
\maketitle


\color{black}
\begin{abstract}
Large language models (LLMs) can expose memorized training sequences under prefix-based extraction: given a prefix from a training example, the model may assign high probability to the original continuation. In deployed systems, however, prefixes are rarely evaluated in isolation. They often appear together with instructions, retrieved documents, or other task-specific context, as in retrieval-augmented generation (RAG). 
This motivates examining whether contextual conditioning mitigates memorization or merely changes the set of memorized samples that become extractable. We investigate this issue through paired item-level measurements of probabilistic suffix extraction.
For each prefix-suffix pair, we score the target suffix under an empty prompt and under retrieved contexts of varying relevance, across three open-weight instruction-tuned models. We find that context does not simply erase memorization. Instead, extractable memorization consists of a context-robust core and a context-sensitive boundary. Many samples that are extractable without context remain extractable under the retrieved context, especially as the prefix length increases. At the same time, context mainly affects marginal samples near the extraction threshold: it suppresses some exposures, but also enables new ones that are missed by prefix-only evaluation. These findings qualify the view that RAG reduces memorization risk. Context can lower aggregate extraction by suppressing boundary cases, yet robustly extractable samples persist, and context-enabled extractability remains security-relevant.~\footnote{The code is available at: \url{https://github.com/alistvt/context-sensitive-memorization}}
\end{abstract}
\color{black}
\section{Introduction}

Language models can memorize parts of their training data, raising privacy and copyright concerns when memorized content can be reproduced at inference time~\cite{carlini2019canaries, henderson2023foundation, mueller2024llms, freeman2024exploring}. The memorized content may be extracted when an adversary supplies a matching prefix of the original sequence~\cite{carlini2021extracting,carlini2022quantifying, hartmann2023sok, satvaty2024undesirable, tirumala2022memorization}. Prefix--suffix extraction attacks measure this behavior directly: given a prefix from a training sequence, the model may assign unusually high probability to the corresponding suffix, enabling verbatim recovery of memorized text.

In many deployed systems, a model is not prompted with an isolated prefix, but with additional text such as retrieved documents, instructions, or task-specific context~\cite{wei2022finetuned, gao2023retrieval}. This is especially relevant for retrieval-augmented generation (RAG), where external documents are attached to the input of the model~\cite{lewis2020retrieval, borgeaud2022improving}. Intuitively, one might expect such context to reduce memorization-based leakage: if the model is grounded in retrieved evidence, it may rely less on parametric memory. This intuition has been explored in recent work suggesting that RAG can reduce leakage~\cite{tgtb}. However, it remains unclear how contextual conditioning changes prefix-based extraction at the level of individual samples.


In this work, we study how extra context affects memorization under the prefix--suffix extraction setting. For each prefix--suffix sample, we measure the probability assigned to the target suffix when the prefix is evaluated alone and when it is evaluated with different types of preceding context. We consider semantically related retrieved context, and irrelevant context, and compare all conditions through paired item-level measurements. This lets us examine not only whether aggregate extraction changes under context, but also how context changes the set of samples that become extractable.
 
Our results show that contextual conditioning does not simply remove memorization. Instead, extractable memorization has a context-robust core of samples that remain extractable regardless of the surrounding context, and a context-sensitive boundary consisting of samples near the extraction threshold whose exposure status flips when context is added. Many core samples remain extractable under retrieved contexts, especially as the prefix length increases. At the same time, context changes the exposure status of marginal samples: some prefix-suffix pairs are suppressed by context, while others become extractable only after context is added. This distinction matters for interpreting claims that retrieval reduces memorization. We find that context can lower aggregate extraction by suppressing boundary cases, but this aggregate reduction should not be interpreted as eliminating memorization risk. Robustly extractable samples remain extractable across contexts, and context-enabled samples are missed by prefix-only evaluation. Thus, contextual conditioning is better understood as reshaping the exposure boundary rather than erasing memorized content.

Taken together, our contributions are:
\begin{itemize}
    \item We formulate prefix--suffix memorization under contextual conditioning and conduct paired item-level comparisons across empty, relevant, and irrelevant contexts.
    \item We show that extractable memorization is composed of a high-confidence core that persists across contexts and weaker extractions that are more sensitive to surrounding context, with robustness increasing as prefix length grows.
    \item We show that aggregate reductions in exposure can hide security-relevant sample-level behavior, including robust exposures that persist across contexts and examples that become extractable only after context is added.
\end{itemize}

\section{Background and Related Work}
\label{sec:background}

\subsection{Memorization in LLMs}
\label{sec:bg-memorization}

Prior work has shown that LLMs can reproduce verbatim training examples under extraction attacks, where an adversary prompts the model and searches generated outputs for memorized training text~\cite{nasr2025scalable, kim2023propile, carlini2021extracting}.

\begin{definition}[Verbatim suffix extraction]
Let $(p,s)$ denote a prefix--suffix pair from a text sequence, where $p$ is given to the model and $s$ is the target continuation. A suffix $s$ is \emph{verbatim extracted} if a model generates $s$ exactly after being prompted with $p$.
\end{definition}

Later studies examined how memorization changes with model scale, data duplication, training dynamics, and language~\cite{carlini2022quantifying,biderman2023emergent, kiyomaru-etal-2024-comprehensive-analysis, tirumala2022memorization, satvaty-etal-2025-memorization}. Our work focuses on verbatim, extractable memorization: given a prefix from a training sequence, we study whether the model assigns high probability to the corresponding suffix.

\subsection{Measuring Memorization}
\label{sec:bg-measuring-memorization}

Memorization is commonly measured through prefix--suffix extraction: given a prefix $p$ from a training sequence, the model is tested on whether it reproduces the corresponding suffix $s$. Much prior work relies on greedy decoding, treating a sample as memorized only when the decoded continuation matches the training suffix~\cite{satvaty2024undesirable}. While intuitive, this binary view can be brittle: a suffix may receive high probability even if it is not produced by a single greedy decode.

Recent work, therefore, measures memorization probabilistically, using the probability assigned to the target suffix rather than only exact decoded matches~\cite{probabilistic}. 

\begin{definition}[Probabilistic extraction]
A suffix $s$ is \emph{probabilistically extractable} if the model assigns high probability to $s$ conditioned on $p$. We measure this by the suffix log-probability
\begin{equation}
    \ell(p,s) = \log P_M(s \mid p).
\end{equation}
Under contextual conditioning, we instead measure
\begin{equation}
    \ell^c(p,s) = \log P_M(s \mid c,p),
\end{equation}
where $c$ is the surrounding context.
\end{definition}

We adopt this view because it gives a continuous per-sample score and supports paired comparisons across contexts. For each prefix--suffix sample, we can measure the suffix probability when the prefix is evaluated alone and when it is evaluated with surrounding context.

Prior work has largely measured extractable memorization from isolated prefixes, where the model is prompted only with a prefix and evaluated on whether it assigns high probability to the original continuation. Closest to our setting is work on privacy in retrieval-augmented generation, which studies whether contextual prompts can reveal sensitive information and finds that retrieval can reduce some forms of leakage. Our setting differs from this line of work in two ways. First, we study verbatim prefix-suffix exposure rather than entity-level disclosure, which lets us evaluate whether the model assigns probability mass to a specific target continuation. Second, our unit of analysis is the individual prefix-suffix pair: we track whether the same sample remains exposed, is suppressed, or becomes newly exposed when context is added. This sample-level view is necessary because aggregate leakage rates can decrease even when a robust subset of exposed samples remains extractable.

\color{black}
\section{Problem Formulation}
\label{sec:problem}

Memorization in language models is commonly evaluated using a prefix--suffix setup: given a prefix from a training sequence, the model is tested on how likely it is to reproduce the corresponding suffix. Most such evaluations implicitly treat memorization as a property of the pair $(p,s)$: if the model assigns high probability to the suffix $s$ after seeing the prefix $p$, then the sample is considered memorized.
In this work, we ask whether this view remains valid when the same prefix is preceded by additional context, i.e. 
retrieved documents. 

\paragraph{Prefix--suffix samples.}
Let $M$ be a language model with parameters $\theta$. We consider a collection
of samples
\begin{equation}
    \mathcal{D} = \{(p_i, s_i)\}_{i=1}^{n},
\end{equation}
where $p_i$ is a prefix and $s_i$ is the corresponding target suffix from the
same source sequence. The standard prefix-only extraction setting evaluates the
model's likelihood of $s_i$ conditioned only on $p_i$.

\paragraph{Contextualized extraction.}
We generalize this setting by introducing a context variable $c_i$. The model is
then evaluated on the likelihood of the same suffix $s_i$ conditioned on both
the context and the prefix:
\begin{equation}
    P_\theta(s_i \mid c_i, p_i).
\end{equation}
The context may be empty, semantically related to the prefix, or intentionally irrelevant. Importantly, the target prefix--suffix pair
$(p_i,s_i)$ remains fixed; only the preceding context changes. This lets us
isolate whether memorization is stable under contextual conditioning.

\paragraph{Context conditions.}
We study the extraction rate of suffix, when the prefix is preceded by three context conditions:

\begin{enumerate}
    \item[\textbf{C.1}] \textbf{Empty context} ($c=\emptyset$): no context is preceded.\label{itm:empty-con}
    \item[\textbf{C.2}] \textbf{Relevant context} ($c=\mathrm{rel}$): the model receives the
    nearest retrieved document from the retrieval collection. \label{itm:rel-con}
    \item[\textbf{C.3}] \textbf{Irrelevant context} ($c=\mathrm{irr}$): the model receives the
    farthest retrieved document, used as a low-relevance context. \label{itm:irr-con}
\end{enumerate}

These context conditions allow us to ask whether memorization is a property of the prefix--suffix pair alone, or whether it depends on the surrounding text in which the prefix is evaluated. In particular, we use them to distinguish between memorized samples that remain extractable across contexts and samples whose exposure changes when contextual information is introduced. We refer to the former as the context-robust core (samples exposed under all context conditions) and to the latter as the context-sensitive boundary (samples whose membership in the exposed set $E_c(\tau)$ depends on c, which we show in Section 5.2 are concentrated near the threshold $\tau$).
\color{black}
\section{Experiments}
\label{sec:setup}

We evaluate memorization under \emph{contextual conditioning}: the same
prefix--suffix sample is scored either in isolation or in the presence of an
additional document context. This setup (more details in Appendix~\ref{app:experimental-setup}) lets us isolate how context
changes prefix-based extraction while keeping the target prefix and suffix
fixed.

Our evaluation has three components: a set of prefix--suffix samples used for
memorization measurement, a document collection from which contextual documents
are selected, and a retrieval procedure that determines which document is placed
before the prefix. For each sample, we compare the empty-context condition
against two contextualized conditions: semantically related
 and irrelevant contexts. All comparisons are paired at the item level:
for a given sample, only the preceding context changes.

\paragraph{Models.}
We evaluate three open-weight instruction-tuned models:
\textbf{Llama-3-8B-Instruct}~\cite{llama3},
\textbf{Falcon-7B-Instruct}~\cite{falcon}, and
\textbf{OLMo-2-1124-7B-Instruct}~\cite{olmo}. The choice of these models is mainly motivated by the transparency of their pretraining data, which makes them suitable for controlled memorization experiments over known data sources.

\paragraph{Memorization Dataset.} We use prefixes sampled from the CommonCrawl
subset of the Pile (\emph{Pile-CC})~\cite{gao2020pile}, which reflects
general web text. 
Specifically, we evaluate on a subset of 50{,}000 prefixes drawn uniformly at random from Pile-CC, without any filtering or pre-selection for memorization. This sampling strategy avoids selection bias toward highly extractable content and follows the standard practice in memorization measurement studies ~\cite{probabilistic, lasy-etal-2025-understanding}, which quantify memorization over sequences sampled directly from the pretraining corpus rather than over curated sets of known-memorized examples. As a result, our exposure sets emerge naturally from the model's behavior on a representative sample of training data, rather than from a regime in which extraction is easiest. We evaluate prefix lengths $\{32, 64, 128\}$ tokens with a fixed suffix length of $32$ tokens.

\paragraph{Retriever and retrieval collection.} We use \textbf{E5 dense retriever}~\cite{wang2024multilingual} for document retrieval. We use this retriever to choose the most relevant and the least relevant documents for each of the prefixes. The
retrieval collection is a mixture of three subsets datasets of the Pile~\cite{gao2020pile}: StackExchange,
PubMed, and Wikipedia subsets were used; for a
total collection of 67{,}655 documents. This mixture is designed to
cover distinct text genres (community Q\&A, biomedical abstracts, and
encyclopedic prose) so that the retrieved context is not topically
biased toward any single domain.


\paragraph{Measurement framework}
\label{sec:measurement-framework}

We adopt the probabilistic extraction approach of \citet{probabilistic}. Rather than measuring memorization only through exact greedy continuation, we estimate how likely the model is to generate the target suffix when conditioned on a prefix, with or without additional context. This allows us to study memorization as a graded likelihood signal instead of a binary generation outcome.

For each sample $i$, we split a text sequence into a prefix $x_i^{\mathrm{pre}}$ and a suffix $x_i^{\mathrm{suf}}$. Given a context condition $c$, we measure the suffix log-probability:
\begin{equation}
    \ell_i^c =
    \log P_\theta \left(
        x_i^{\mathrm{suf}}
        \mid c, x_i^{\mathrm{pre}}
    \right),
\end{equation}
where $P_\theta$ is the language model. The context $c$ may be empty, semantically related, or irrelevant. We denote the corresponding log-probabilities as:
\begin{equation}
    \ell_i^{\emptyset}, \quad
    \ell_i^{\mathrm{rel}}, \quad
    \ell_i^{\mathrm{irr}},
\end{equation}
where $\ell_i^{\emptyset}$ is the suffix log-probability under the empty-context condition, $\ell_i^{\mathrm{rel}}$ under the nearest retrieved context, and $\ell_i^{\mathrm{irr}}$ under the farthest retrieved context.

To analyze whether a sample is extractable under a given condition, we define an exposed set at threshold $\tau$:
\begin{equation}
    E_c(\tau) = \{ i : \ell_i^c > \tau \}.
\end{equation}

This formulation lets us compare memorization across contexts at both the aggregate and sample levels. At the aggregate level, we compare the distributions of $\ell_i^c$ across context conditions. At the sample level, we analyze how membership in $E_c(\tau)$ changes when context is added. In particular, we distinguish between samples that remain exposed across contexts, samples that are suppressed by context, and samples that become exposed only under contextual conditioning.
\color{black}
\section{Results}
\label{sec:results}

\begin{figure*}[!t]
    \centering
    \includegraphics[width=\textwidth]{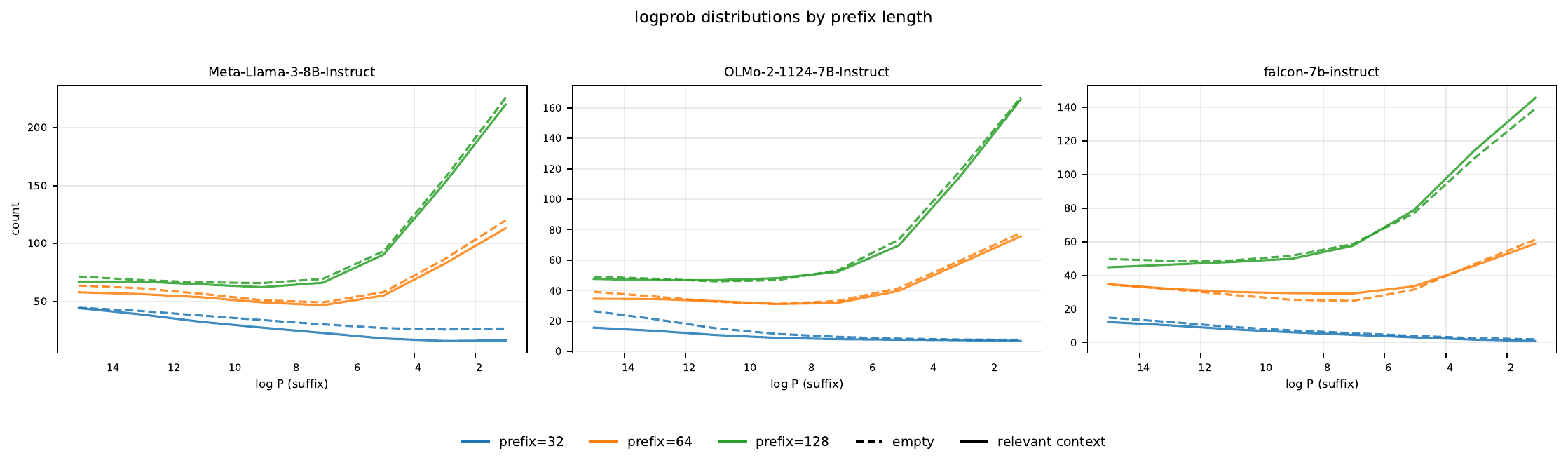}
    \caption{
    Suffix log-probability distributions by prefix length. Dashed lines show empty-context prompting, while solid lines show prompting with retrieved context. Across all models, longer prefixes substantially increase suffix likelihood. The gap between empty and contextualized prompting is largest for shorter prefixes and becomes smaller as prefix length increases, suggesting that longer prefixes make memorization more context-independent.
    }
    \label{fig:logprob-by-prefix}
\end{figure*}

\begin{figure*}[!t]
    \centering
    \includegraphics[width=\textwidth]{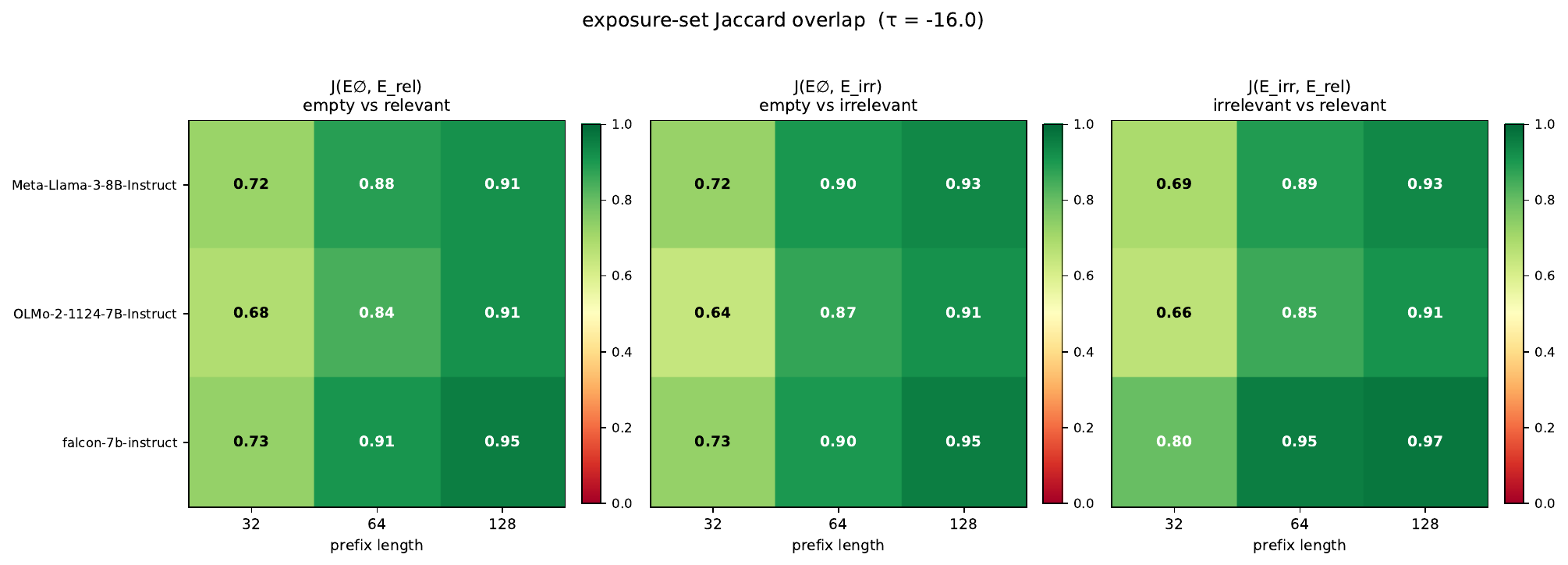}
    \caption{
    Jaccard overlap between exposed sets at $\tau=-16$. Each cell reports the overlap between two exposure sets for a given model and prefix length. Exposure-set overlap increases sharply with prefix length across all models and context comparisons. This indicates that longer prefixes make the set of extractable samples more stable across contexts.
    }
    \label{fig:jaccard-overlap}
\end{figure*}

The results support a consistent picture: retrieved context does not erase prefix-based memorization, but reshapes the exposure boundary. We first show that prefix length dominates contextual variation: longer prefixes make suffix likelihoods and exposed sets increasingly stable across contexts. We then show that this stability comes from a robust core of samples exposed under multiple contexts, while context-induced changes are concentrated near the exposure threshold. Finally, we separate the generic effect of adding context from the semantic effect of retrieving a related document.

We report four observations, which we describe in the following subsections. 

Throughout this section, we use the measurement framework defined in Section~\ref{sec:measurement-framework}. Recall that $\ell_i^{\emptyset}$, $\ell_i^{\mathrm{rel}}$, and $\ell_i^{\mathrm{irr}}$ denote the suffix log-probability of sample $i$ under empty context, relevant retrieved context, and irrelevant retrieved context, respectively. We define the exposed set under condition $c$ as
\begin{equation}
    E_c(\tau) = \{i : \ell_i^c > \tau\}.
\end{equation}
We use $\tau=-16$ in the main analysis\footnote{Equivalent to 0.5 per-token drop of logprob for 32 tokens of suffix.} and report threshold-sensitivity results in the appendix \ref{app:threshold-sensitivity}.

\subsection{Observation 1: Longer prefixes make memorization increasingly context-independent}
\label{sec:res-prefix-context-independent}

We first ask whether contextual conditioning changes prefix-based memorization measurements, and whether this effect depends on prefix length. Figure~\ref{fig:logprob-by-prefix} shows the distribution of suffix log-probability scores for different prefix lengths under empty and retrieved-context prompting. Across all three models, increasing the prefix length from 32 to 64 and 128 tokens shifts the distribution toward higher suffix likelihoods. This confirms that prefix length is a dominant factor in extraction: longer prefixes provide a stronger conditioning signal and make the memorized continuation more likely.

Importantly, the difference between empty-context and contextualized prompting becomes smaller as the prefix length increases. For short prefixes, contextual conditioning can noticeably shift the suffix likelihood distribution. For longer prefixes, the empty-context and contextualized curves become much closer. This suggests that longer prefixes make extraction increasingly \emph{context-independent}: the model relies more on the prefix itself and less on the preceding context.

We quantify this stability using Jaccard overlap between exposed sets:
\begin{equation}
    J(E_a,E_b) = \frac{|E_a \cap E_b|}{|E_a \cup E_b|}.
\end{equation}
Figure~\ref{fig:jaccard-overlap} shows that exposed-set overlap increases consistently with prefix length for all three models. This trend holds for empty versus relevant/irrelevant context and irrelevant versus relevant context.

This provides a direct answer to our motivating question: memorization is not uniformly context-independent, but it becomes increasingly context-independent as the prefix becomes more informative. Short-prefix extraction is more context-sensitive, while long-prefix extraction is more stable across contextual conditions.

\begin{figure*}[!t]
    \centering
    \includegraphics[width=\textwidth]{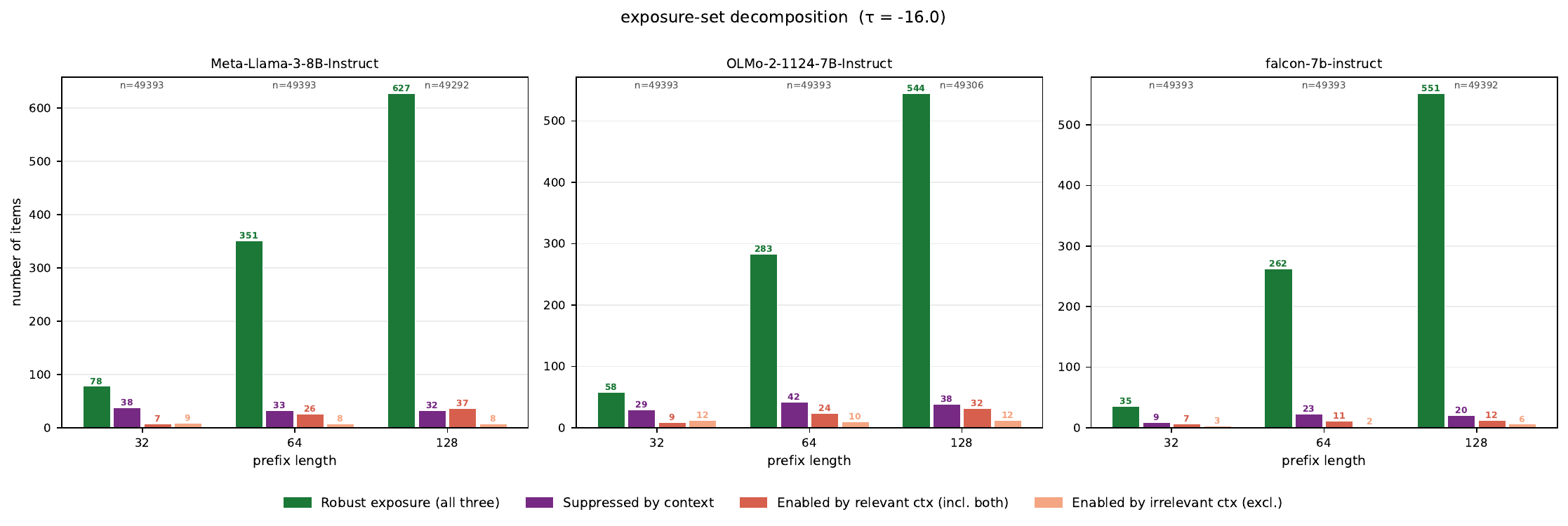}
    \caption{
    Exposure-set decomposition at $\tau=-16$. Green bars show samples exposed under all three conditions. Purple bars show samples exposed without context but suppressed by context. Red and peach bars show samples that become exposed only under relevant or irrelevant context. Robust exposure dominates across models and grows substantially with prefix length, indicating a context-robust memorization core.
    }
    \label{fig:exposure-decomposition}
\end{figure*}

\subsection{Observation 2: Strongly exposed samples persist across contexts, while weaker exposures are context-sensitive}
\label{sec:res-robust-core-boundary}

The previous result shows that exposure sets become more stable with longer prefixes. We now ask whether this stability comes from a shared robust core of memorized samples or whether the context completely reshuffles which samples are exposed.

Figure~\ref{fig:exposure-decomposition} decomposes exposed samples into four groups: samples exposed under all three conditions, samples exposed without context but suppressed by context, samples enabled by relevant context, and samples enabled only by irrelevant context. The dominant category is robust exposure which consistently increases with the prefix length for all three models.

This indicates that context does not erase memorization. A large fraction of exposed samples remain extractable under empty, relevant, and irrelevant contexts. The robust core becomes especially pronounced as the prefix length increases, consistent with the Jaccard-overlap results in Figure~\ref{fig:jaccard-overlap}.\footnote{See Table~\ref{tab:jaccard-overlap} in Appendix \ref{app:exposure-set-analysis} for additional results.}

\begin{table*}[t]
\centering
\small
\begin{tabular}{llrrrrrr}
\toprule
Model & Prefix & $n$ & Mean $G$ & Median $G$ & Mean $S$ & Median $S$ & \% $S<0$ \\
\midrule
\multirow{3}{*}{Llama-3-8B-Instruct}
& 32  & 318 & 0.52  & -0.34 & -3.35 & -1.24 & 70.1 \\
& 64  & 593 & 0.52  & -0.06 & -1.11 & -0.12 & 59.7 \\
& 128 & 858 & 0.05  & -0.03 & -0.32 & -0.04 & 57.5 \\
\midrule
\multirow{3}{*}{OLMo-2-1124-7B-Instruct}
& 32  & 131 & -0.42 & -0.23 & -2.05 & -0.75 & 62.6 \\
& 64  & 390 & -0.32 & -0.06 & -0.04 & 0.06  & 47.2 \\
& 128 & 645 & 0.14  & -0.01 & -0.30 & 0.01  & 47.3 \\
\midrule
\multirow{3}{*}{Falcon-7B-Instruct}
& 32  & 69  & -0.94 & -1.02 & -1.70 & -0.64 & 60.9 \\
& 64  & 321 & 0.56  & 0.10  & -0.23 & -0.13 & 56.1 \\
& 128 & 617 & 0.42  & 0.19  & -0.16 & -0.06 & 54.3 \\
\bottomrule
\end{tabular}
\caption{
Generic and semantic context effects for exposed or boundary samples. We report the number of boundary samples $n$, the generic context effect $G=\ell^{\mathrm{irr}}-\ell^{\emptyset}$, and the semantic effect $S=\ell^{\mathrm{rel}}-\ell^{\mathrm{irr}}$. The semantic effect is generally negative, especially for shorter prefixes, indicating that relevant context tends to suppress extraction relative to irrelevant context near the exposure boundary. The generic effect is more model-dependent.
}
\label{tab:semantic-generic}
\end{table*}

Meanwhile, Figure~\ref{fig:exposure-decomposition} also shows that context-sensitive samples are non-negligible. Some samples are removed from the exposed set when context is added, while others become exposed only under contextual conditioning. To understand where these changes occur, we examine the distance of each sample from the empty-context exposure threshold:
\begin{equation}
    m_i = |\ell_i^{\emptyset} - \tau|.
\end{equation}
We then compute how often samples flip exposure status after context is added.

We find that context-induced flips are highly concentrated near the exposure boundary (details in Figure~\ref{fig:margin-flip} in Appendix~\ref{app:flip-analysis}). Samples in the closest bin to the threshold have the highest flip rates, while samples far from the threshold almost never flip. This pattern appears across models and prefix lengths. As the prefix length increases, the relevant and irrelevant context curves also become more similar, especially near the boundary, indicating that longer prefixes reduce both the number of flips and the dependence of flips on the type of retrieved context. 

Together, these results suggest that memorization 
of strongly exposed samples remains stable across contexts, whereas less memorized samples are more sensitive to preceding context.

\subsection{Observation 3: Context mostly suppresses, but context-enabled extraction remains security-relevant}
\label{sec:res-suppression-enabling}

The exposure decomposition also allows us to revisit prior claims that retrieval or contextualization reduces memorization. Our results are consistent with this claim at the aggregate level: context more often suppresses exposure than enables it. In Figure~\ref{fig:exposure-decomposition}, the suppressed category is generally larger than the context-enabled categories.

However, the security interpretation is more subtle. A reduction in average exposure does not imply that memorization risk has been removed. First, many samples remain robustly exposed across all contexts, as shown by the green bars in Figure~\ref{fig:exposure-decomposition}. Second, context-enabled exposure is consistently nonzero. That is, some samples that are not extractable under empty-context prompting become extractable when preceded by some context.\footnote{Appendix~\ref{app:examples} shows several examples of such cases.}

This matters for evaluating RAG and other context-augmented systems. 
Our results show that contextualization can act as a filter: it suppresses some marginal examples, preserves a robust core, and occasionally enables new exposure.
Context changes the exposure boundary. This reconciles the observation that RAG can reduce aggregate memorization with the persistence of robustly extractable samples.

\subsection{Observation 4: Relevant context adds a semantic effect, but generic context also matters}
\label{sec:res-semantic-generic}

Finally, we ask whether the effect of context is caused by semantic relevance or by the generic presence of additional context. We decompose the effect into two quantities:
\begin{align}
    G_i &= \ell_i^{\mathrm{irr}} - \ell_i^{\emptyset}, \\
    S_i &= \ell_i^{\mathrm{rel}} - \ell_i^{\mathrm{irr}}.
\end{align}
Here, $G_i$ captures the generic effect of adding irrelevant context, while $S_i$ captures the additional semantic effect of replacing irrelevant context with relevant retrieved context.

Table~\ref{tab:semantic-generic} (also shown as violin plots in Figure~\ref{fig:semantic-generic} in  Appendix~\ref{sub:semantic-relevance}) shows this decomposition for exposed samples (if a sample exceeds the threshold in any of the three prefix scenarios, it is included). The semantic effect $S$ is often negative, especially for shorter prefixes. This suggests that relevant retrieved context often suppresses extraction relative to irrelevant context.

The generic effect $G$, however, is more model-dependent. In some settings, irrelevant context suppresses extraction; in others, it has little effect or even increases likelihood. This indicates that contextual effects are not purely semantic. Adding context itself can perturb the conditional distribution, while semantic relevance can add an additional suppressive bias for boundary samples.


\paragraph{Summary.}
Our results show that memorization is neither fully context-independent nor fully context-dependent. Instead, exposed memorization consists of a high-confidence core that persists across contexts and a weaker set of extractions that is more sensitive to surrounding text. Longer prefixes enlarge the persistent core, while context mainly affects lower-confidence samples. Thus, contextualization can reduce aggregate extraction by suppressing weaker extractions, but it does not eliminate robustly extractable memorization and can also enable new exposure. Item-level analysis of suffix log-probabilities under relevant and irrelevant context is available in Appendix~\ref{app:scatter-rel-irr}.

\subsection{Ablation analysis}

\paragraph{Choice of memorization threshold}
The probabilistic extraction framework assigns each item a continuous suffix
log-probability score rather than a binary memorized/non-memorized label. For
threshold-based analyses, we use $\tau=-16$ as the exposure threshold. This
choice is intentionally less strict than only counting very high-confidence
extractions, allowing us to capture weaker memorized samples that may still be
relevant for contextual exposure. 
Appendix~\ref{app:threshold-sensitivity} gives a security interpretation of this threshold in terms of adversary sampling budgets and shows that our conclusions are stable across thresholds spanning five orders of magnitude of extraction probability.

\paragraph{Instruction-Framed Contextualization}

\begin{figure}[t]
    \centering
    \includegraphics[width=\columnwidth]{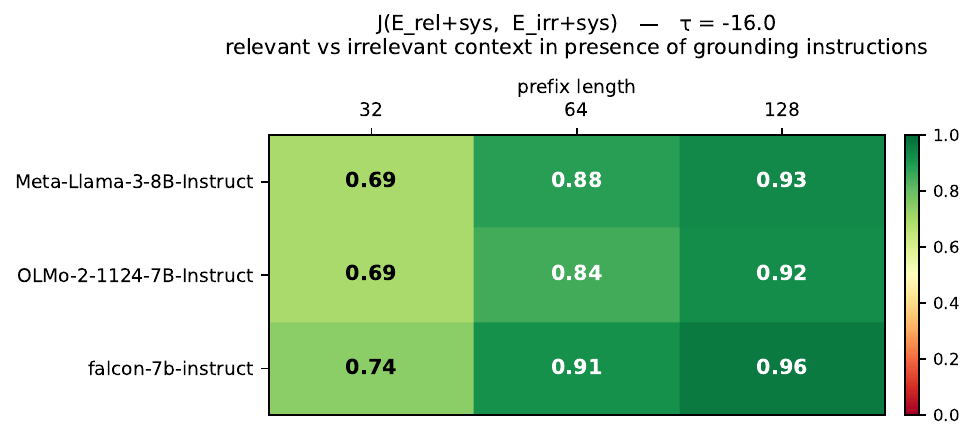}
    \caption{
    Jaccard overlap between exposed sets under relevant and irrelevant retrieved context in a fixed instruction-framed setting. Exposure-set overlap increases with prefix length across all models.
    }
    \label{fig:sys_rel_rand_jaccard}
\end{figure}

We repeat part of our analysis under an instruction-framed prompt. 
We find that the Jaccard overlap between exposed sets under relevant and irrelevant retrieved contexts is already moderate for prefix length 32 and increases substantially with prefix length in all three models (Figure~\ref{fig:sys_rel_rand_jaccard}). This suggests that under instruction-framed prompting, retrieval choice still affects some boundary cases, but most exposed samples overlap between context types (see Appendix \ref{app:instruction_framed} for more quantitative results).

\paragraph{Effect of Chat Templates}

In our experiments we consider working without the chat templates, since we try to mainly measure the effect of context preceding prefix. 
We notice that experimenting with a setting where we place the prefix inside a full chat template requires a somewhat different evaluation setup and introduces a different measurement regime: For the sequence $(p,s)$ that appeared in training, the model is no longer asked to continue $p$ with $s$, but instead to generate $s$ after an assistant tag. We find that this has a substantial effect: across all models and prefix lengths, adding the chat template substantially lowers measured memorization. The implication is that chat-template formatting makes the results harder to interpret by mixing the effect of preceding context with a different response-formatting regime. More details are in Appendix~\ref{app:chat-template}.
\section{Discussion}
\label{sec:discussion}
\paragraph{Applicability to real-world systems}
Our setup deliberately isolates the effect of retrieved context on
memorization by controlling the chat template, the retriever, and the
collection. This isolation does not narrow the applicability of the
findings: the core mechanism we study -- adding a retrieved document to
the prompt of an instruction-tuned model -- is the same operation that
real-world RAG systems perform. Additional components of production
pipelines (prompt rewriting, multi-turn dialogue, output filtering,
post-hoc safety classifiers) operate around this core operation rather
than replacing it. The conclusion that memorized items remain
extractable under retrieval, and that retrieval surfaces items the LLM
alone does not, therefore carries over to systems built on top of the
same retrieval-and-prompting pattern.

\paragraph{Memorization as a proxy for leakage}
We measure targeted (prefix-suffix) memorization instead of leakage, because leakage itself does not afford a
fair comparison between conditions: leakage can occur inadvertently
during benign use, and it can be elicited by deliberate extraction
attacks, so two systems with identical memorization profiles may
yield very different leakage rates depending on the attack regime
they are tested under. The probabilistic memorization metric
factors this confound out: it captures the underlying tendency of
the model to assign non-zero probabilities to a memorized continuation. This probabilistic memorization effect can lead to both inadvertent and adversarial leakage and is therefore a neutral measurement.
Items whose memorization is preserved in both plain-LLM and RAG
conditions therefore remain equally leakable in the operational
sense, even if the specific leakage rate observed in any given
deployment will depend on the specific attack model.

\section{Conclusion}
\label{sec:conclusion}

We studied whether prefix-based memorization remains unchanged when the prefix is evaluated with surrounding context. Our results show that contextual conditioning does not simply remove memorization. 
Across three instruction-tuned models, many extractable samples remain exposed under related and irrelevant contexts, especially as prefix length increases. Context-induced changes are concentrated near the exposure threshold, while samples far from the threshold are largely stable. 

These findings nuance the claim that retrieval reduces memorization. Context can reduce aggregate extraction by suppressing boundary cases, but robustly exposed samples remain, and some samples become extractable only after context is added. 
Thus, context should be viewed less as a memorization shield and more as a filter that reshapes which samples are exposed. Future work should extend these analyses to additional datasets and experimental settings.

\section*{Limitations}
\label{sec:limitations}
Our experiments focus on open-weight models from three families, Llama-3, OLMO-2, and Falcon, at approximately the 10B-parameter scale, and use a randomly sampled subset of the CommonCrawl portion of Pile-CC as prefix data. Although we do not evaluate the largest open-weight models or closed APIs, the models studied here are already large enough to exhibit substantial memorization behavior and to support meaningful analysis of extraction dynamics. Evaluating larger models could further clarify how absolute memorization rates change with scale, but larger models are not necessary for establishing our main qualitative finding: memorized continuations remain largely context-independent across diverse prompting conditions. Similarly, domains outside Pile-CC, such as code, biomedical text, or copyrighted creative writing, may show different absolute memorization rates. Nevertheless, CommonCrawl-derived text is widely used in memorization studies, spans diverse sources and writing styles, and appears across the pre-training corpora of the model families we evaluate. This makes it a suitable basis for cross-model comparison, and the consistency of our findings across model families suggests that domain variation is unlikely to overturn the central conclusion.

Our probabilistic extraction scores are computed with temperature $T=1$ and top-$p=1$. This setting preserves the model's original probability distribution and avoids truncating suffix tokens, ensuring that every sample receives a finite and comparable memorization score. We also report retrieval-augmented results using the E5 dense retriever under both nearest and farthest retrieval, reflecting common real-world RAG settings. Preliminary experiments suggest that replacing E5 with other lexical or dense retrievers has only a limited effect on memorization, but a broader retriever sweep remains an important direction for future work. Other decoding or retrieval choices may affect surface-level generations, yet they are unlikely to change the core probabilistic comparison studied here.

Finally, we evaluate multiple prefix lengths, $\{32, 64, 128\}$, with a fixed suffix length of $32$. Longer suffixes may lead to different quantitative patterns because token-level probability gaps can compound over longer continuations. A systematic sweep over suffix length is therefore left to future work.

\section*{Ethical Considerations}
Our prefixes are drawn from a publicly available corpus (Pile-CC); we do not target private or proprietary training data, closed API models, or deployed systems. Our quantitative results are reported as aggregate metrics. For qualitative analysis we include a limited number of illustrative examples in the appendix, consisting of prefix–suffix pairs and their retrieved contexts drawn from Pile-CC; these are already public web text, and we do not release the full set of per-example extraction outputs. We release our code to support reproducibility. Our findings argue against treating RAG as a defense for memorization-based leakage and motivate defense efforts targeted at parametric memory itself.

\color{black}

\section*{Acknowledgements}
This publication is part of the project LESSEN with project number NWA.1389.20.183 of the research program NWA ORC 2020/21 which is (partly) financed by the Dutch Research Council (NWO).

We would like to thank KPN for hosting part of this research. Part of this work was carried out by Narjes during a research internship at KPN, and we thank Gianluigi Bardelloni for his supervision and support.
\bibliography{references}

\newpage

\appendix

\section{Threshold Choice and Sensitivity Analysis}
\label{app:threshold-sensitivity}
Our main analysis defines exposed samples using a fixed threshold $\tau = -16$. For a 32-token suffix, this corresponds to an average per-token log-probability of $-0.5$, i.e., a mean per-token probability of $e^{-0.5} \approx 0.61$, and a total suffix probability of $e^{-16} \approx 1.1 \times 10^{-7}$. In the discoverable-extraction sense of \citet{probabilistic}, an adversary sampling at temperature 1 recovers the exact suffix of a sample at this threshold with probability $\geq 0.5$ within on the order of $10^6$--$10^7$ generation attempts---a realistic budget for automated extraction attacks. We deliberately choose this permissive operating point rather than one that counts only near-deterministic extractions, because our analysis targets the exposure \emph{boundary}, and a stricter threshold would truncate precisely the marginal region where context-induced changes concentrate.

\begin{figure*}[t]
    \centering
    \includegraphics[width=\textwidth]{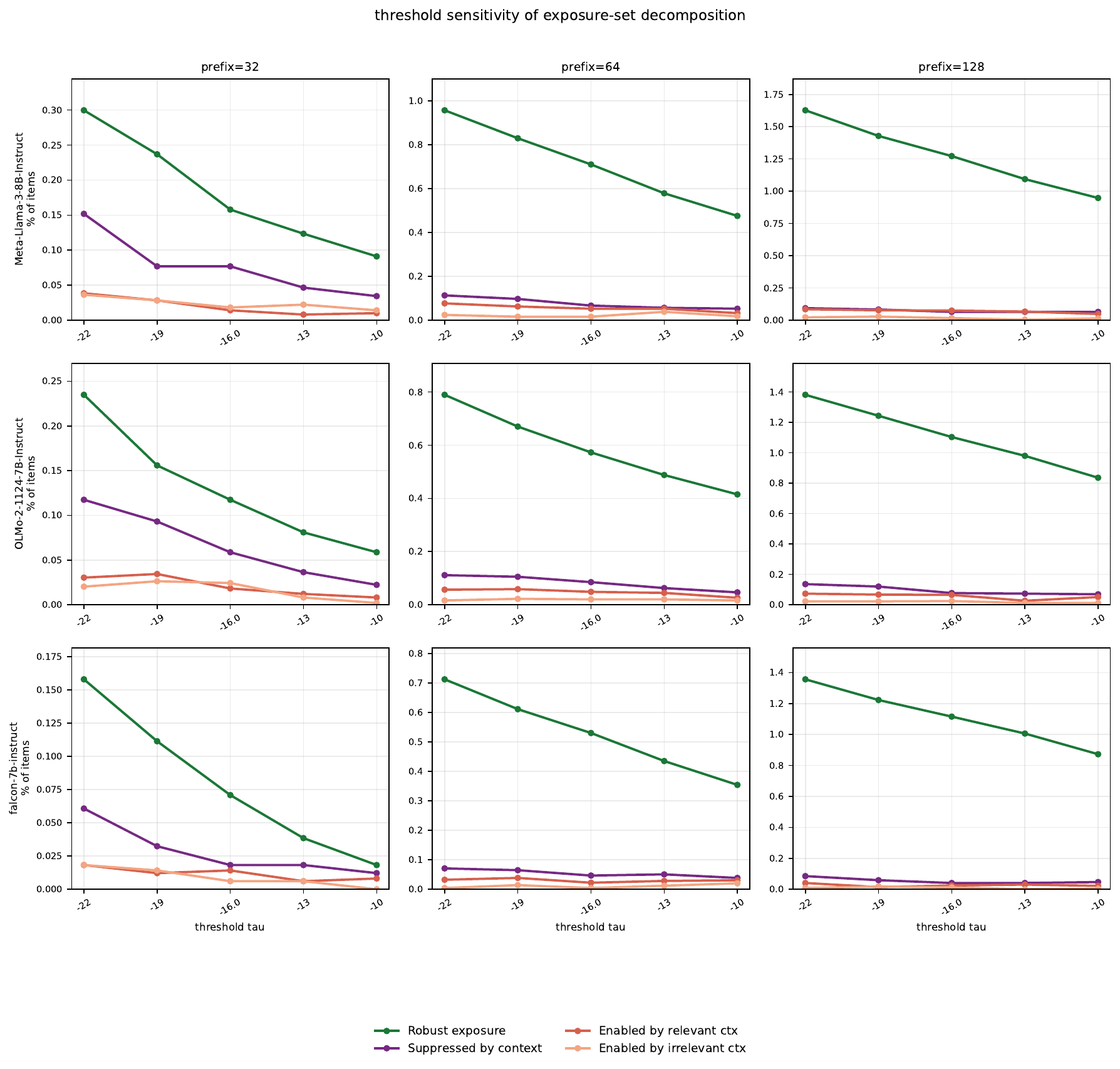}
    \caption{
    Threshold sensitivity of the exposure-set decomposition. We vary the exposure threshold $\tau$ and report the percentage of samples in each category. Robust exposure remains the dominant category across thresholds, especially for longer prefixes. Suppressed and context-enabled samples remain smaller but persist across threshold choices, showing that the exposure-set decomposition is not an artifact of a single cutoff.
    }
    \label{fig:threshold-sensitivity-exposure}
\end{figure*}

To test whether our conclusions depend on this particular cutoff, we repeat the exposure-set decomposition across multiple thresholds. Specifically, for each context condition $c$, we define the exposed set as
\begin{equation}
    E_c(\tau) = \{ i : \ell_i^c > \tau \},
\end{equation}
and vary $\tau$ over the range $[-22, -10]$, corresponding to adversary budgets from ${\sim}10^9$ down to ${\sim}10^4$ generation attempts, i.e., from extremely powerful to cheap extraction attacks.
Figure~\ref{fig:threshold-sensitivity-exposure} shows that the qualitative structure of the exposed sets is stable across thresholds. As expected, stricter thresholds reduce the absolute fraction of exposed samples. However, the relative pattern remains consistent: robust exposure is the dominant category, especially for longer prefixes, while suppressed and context-enabled samples remain smaller but nonzero. This indicates that our main conclusions are not artifacts of a single threshold choice.
Importantly, the context-enabled categories persist across thresholds. Although they are much smaller than the robust exposed set, their presence shows that context-induced exposure is not caused solely by an arbitrary cutoff.

\section{Flip Analysis}
\label{app:flip-analysis}

In the main text, we show that contextual conditioning can change whether a sample is classified as exposed. Here, we further analyze where these flips occur relative to the exposure threshold. For each sample, we compute its distance from the empty-context threshold,
\begin{equation}
    m_i = |\ell_i^{\emptyset} - \tau|,
\end{equation}
and measure how often the sample changes exposure status after adding context.

\begin{figure*}[t]
    \centering
    \includegraphics[width=\textwidth]{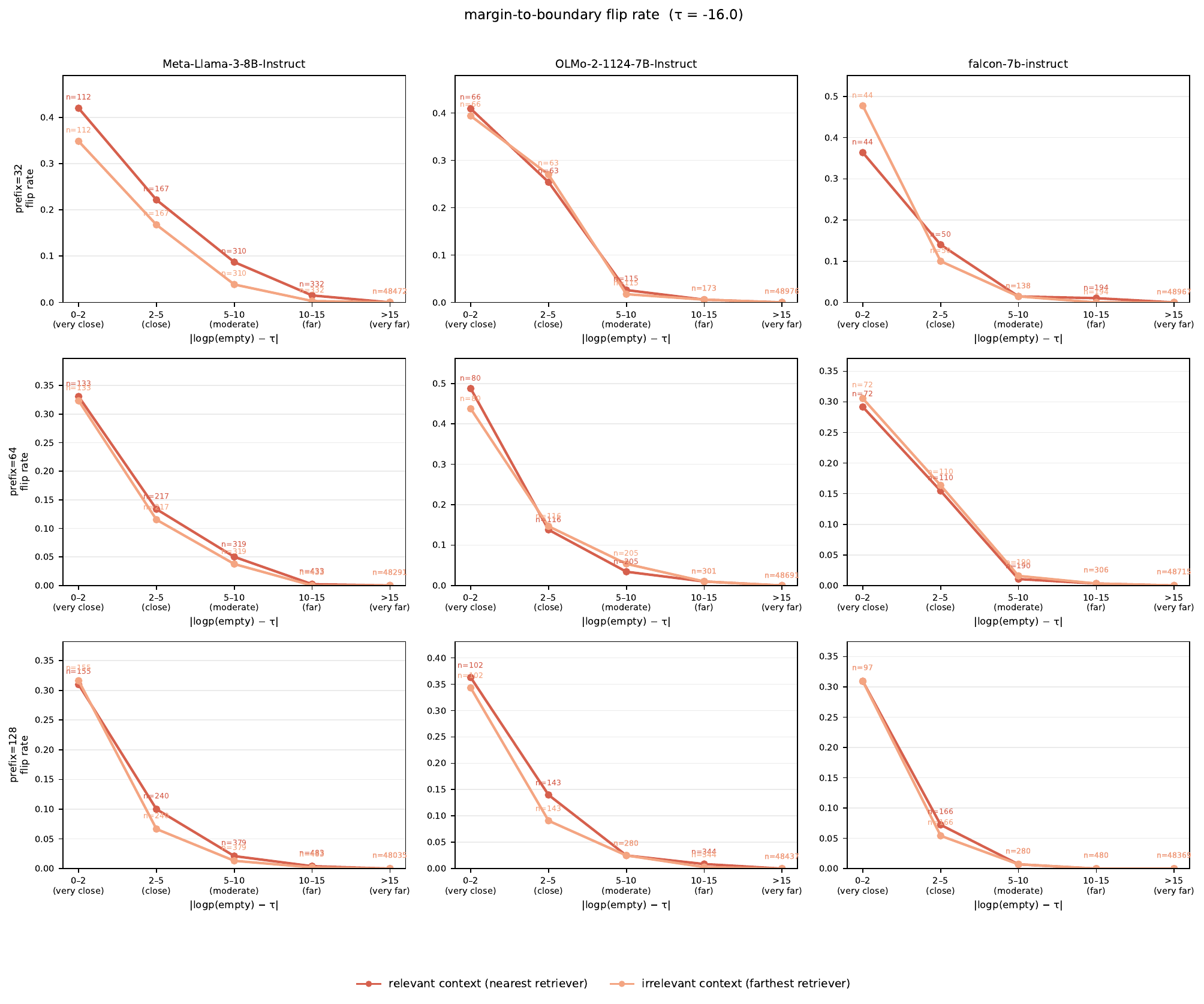}
    \caption{
    Context-induced flip rate as a function of distance from the empty-context exposure threshold. The flip rate is highest for samples close to the threshold and approaches zero for samples far from the threshold. This indicates that context-induced changes are boundary-local: context primarily affects marginal samples rather than robustly exposed or strongly non-exposed samples.
    }
    \label{fig:margin-flip}
\end{figure*}

Figure~\ref{fig:margin-flip} shows that context-induced flips are concentrated near the threshold. Samples close to the boundary have the highest flip rates, while samples far from the boundary almost never flip. This supports the interpretation that context primarily affects marginal samples rather than strongly exposed or strongly non-exposed samples.

\section{Exposure Set Analysis}
\label{app:exposure-set-analysis}

In addition to the exposure-set decomposition in the main text, we report the
pairwise overlap between exposed sets under different context conditions. For
each condition $c$, we define the exposed set as
\begin{equation}
    E_c(\tau) = \{i : \ell_i^c > \tau\}.
\end{equation}
We then compute the Jaccard overlap between exposed sets,
\begin{equation}
    J(E_a,E_b) = \frac{|E_a \cap E_b|}{|E_a \cup E_b|}.
\end{equation}

Table~\ref{tab:jaccard-overlap} reports the overlap between empty, relevant,
and irrelevant context at $\tau=-16$. Across all models, the overlap increases
substantially with prefix length. This supports the main result that longer
prefixes make the set of exposed samples more consistent across contextual
conditions.

\begin{table*}[t]
\centering
\small
\begin{tabular}{llrrrrrr}
\toprule
Model & Prefix & $J(E_{\emptyset},E_{\mathrm{rel}})$ & $J(E_{\emptyset},E_{\mathrm{irr}})$ & $J(E_{\mathrm{irr}},E_{\mathrm{rel}})$ & $|E_{\emptyset}|$ & $|E_{\mathrm{rel}}|$ & $|E_{\mathrm{irr}}|$ \\
\midrule
\multirow{3}{*}{Meta-Llama-3-8B-Instruct}
& 32  & 0.715 & 0.721 & 0.689 & 116 & 95  & 106 \\
& 64  & 0.880 & 0.901 & 0.890 & 384 & 387 & 384 \\
& 128 & 0.912 & 0.933 & 0.932 & 659 & 672 & 665 \\
\midrule
\multirow{3}{*}{OLMo-2-1124-7B-Instruct}
& 32  & 0.677 & 0.637 & 0.656 & 87  & 74  & 80  \\
& 64  & 0.840 & 0.867 & 0.854 & 325 & 317 & 319 \\
& 128 & 0.912 & 0.911 & 0.909 & 582 & 592 & 580 \\
\midrule
\multirow{3}{*}{Falcon-7B-Instruct}
& 32  & 0.725 & 0.725 & 0.796 & 44  & 44  & 44  \\
& 64  & 0.905 & 0.898 & 0.951 & 285 & 279 & 275 \\
& 128 & 0.952 & 0.950 & 0.967 & 571 & 567 & 568 \\
\bottomrule
\end{tabular}
\caption{
Exposure-set Jaccard overlap at $\tau=-16$. $E_{\emptyset}$,
$E_{\mathrm{rel}}$, and $E_{\mathrm{irr}}$ denote the exposed sets under empty,
relevant, and irrelevant context, respectively. Across all models, exposed-set
overlap increases with prefix length, indicating that longer prefixes make
exposure less sensitive to contextual variation.
}
\label{tab:jaccard-overlap}
\end{table*}

\section{Instruction-Framed Contextualization}
\label{app:instruction_framed}

As an additional robustness check, we repeat part of our analysis under a fixed instruction-framed prompt. Because this setting changes both the prompting format and the role structure, we do not interpret it as an isolated measurement of system-prompt effects. Instead, we use it to test whether our main conclusions about exposure stability continue to hold in a chat-style setting.

Figure~\ref{fig:sys_rel_rand_jaccard} shows the Jaccard overlap between exposed sets under relevant and irrelevant retrieved contexts, with the instruction framing held fixed. The overlap is already moderate for prefix length 32 and increases substantially with prefix length across all three models. For example, the overlap rises from 0.69 to 0.93 for Llama, from 0.69 to 0.92 for OLMo, and from 0.74 to 0.96 for Falcon as the prefix length increases from 32 to 128 (Table \ref{tab:sys-rel-rand-jaccard}). This mirrors the main-paper trend: longer prefixes make exposure increasingly stable across contextual conditions.

\begin{figure*}[t]
    \centering
    \includegraphics[width=\linewidth]{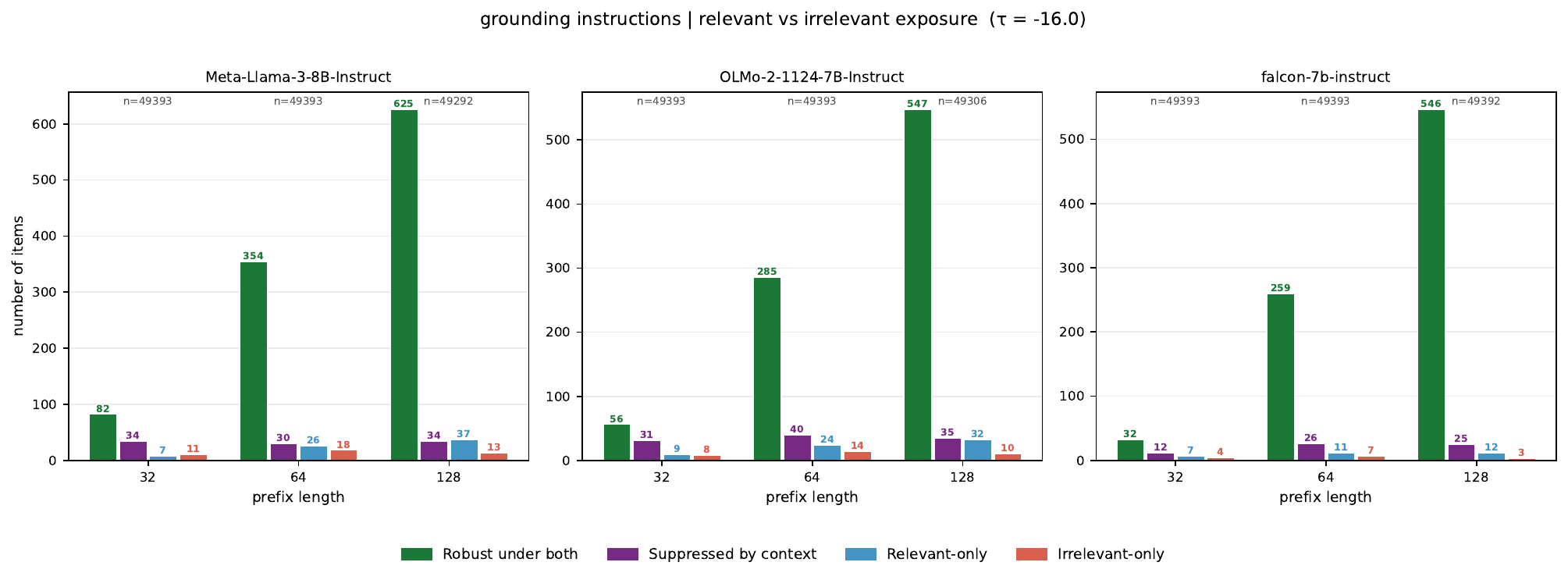}
    \caption{
    Exposure decomposition for relevant versus irrelevant retrieval under a fixed instruction-framed setting. Most exposed samples are shared between the two retrieval conditions, and the shared robust set grows with prefix length.
    }
    \label{fig:sys_rel_rand_decomp}
\end{figure*}

\begin{figure*}[t]
    \centering
    \includegraphics[width=\linewidth]{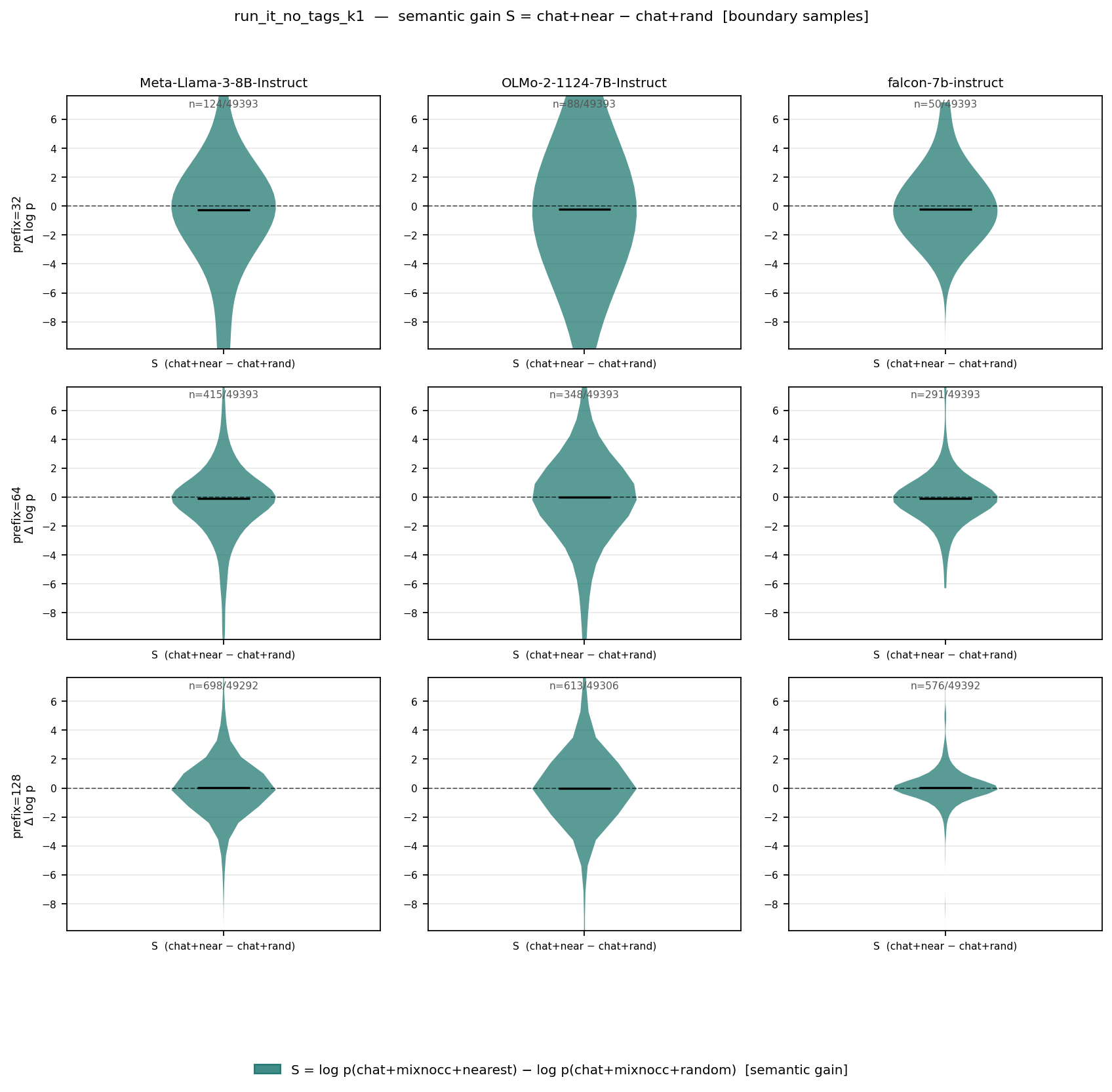}
    \caption{
    Semantic with system prompt
    }
    \label{fig:sys_rel_rand_decomp_1}
\end{figure*}

Figure~\ref{fig:sys_rel_rand_decomp} further decomposes the exposed sets into samples shared by relevant and irrelevant retrieval, samples exposed only under relevant retrieval, and samples exposed only under irrelevant retrieval. In all models, the shared category dominates and grows substantially with prefix length, while the relevant-only and irrelevant-only categories remain comparatively small. This suggests that under instruction-framed prompting, retrieval choice still affects some boundary cases, but most exposed samples belong to a shared robust core. The corresponding semantic effect S under grounding instructions is shown in Figure \ref{fig:sys_rel_rand_decomp_1} and summarized in Table \ref{tab:semantic-boundary-effect}.

Overall, these results reinforce the main conclusion of the paper: memorization is characterized by a robust exposed core and a context-sensitive boundary, and the robust core becomes increasingly dominant as the prefix length grows.

\begin{table*}[t]
\centering
\small
\begin{tabular}{llrrrr}
\toprule
Model & Prefix & $J(E_{\mathrm{rel}},E_{\mathrm{irr}})$ & $|E_{\mathrm{rel}}|$ & $|E_{\mathrm{irr}}|$ & $n_{\mathrm{total}}$ \\
\midrule
\multirow{3}{*}{Meta-Llama-3-8B-Instruct}
& 32  & 0.694 & 95  & 115 & 49393 \\
& 64  & 0.882 & 387 & 394 & 49393 \\
& 128 & 0.927 & 672 & 673 & 49292 \\
\midrule
\multirow{3}{*}{OLMo-2-1124-7B-Instruct}
& 32  & 0.693 & 74  & 75  & 49393 \\
& 64  & 0.845 & 317 & 325 & 49393 \\
& 128 & 0.920 & 592 & 585 & 49306 \\
\midrule
\multirow{3}{*}{Falcon-7B-Instruct}
& 32  & 0.740 & 44  & 43  & 49393 \\
& 64  & 0.911 & 279 & 277 & 49393 \\
& 128 & 0.958 & 567 & 561 & 49392 \\
\bottomrule
\end{tabular}
\caption{
Jaccard overlap between exposed sets under relevant and irrelevant context in the instruction-framed setting. $E_{\mathrm{rel}}$ and $E_{\mathrm{irr}}$ denote the exposed sets under relevant and irrelevant retrieved contexts, respectively. The overlap increases with prefix length across all models, indicating that longer prefixes make exposure increasingly stable across retrieved-context choices.
}
\label{tab:sys-rel-rand-jaccard}
\end{table*}

\begin{table*}[t]
\centering
\small
\begin{tabular}{llrrrrr}
\toprule
Model & Prefix & $n_{\mathrm{total}}$ & $n_{\mathrm{boundary}}$ & Mean $S$ & Median $S$ & \% $S<0$ \\
\midrule
\multirow{3}{*}{Meta-Llama-3-8B-Instruct}
& 32  & 49393 & 124 & -0.998 & -0.241 & 56.5 \\
& 64  & 49393 & 415 & -0.496 & -0.108 & 54.9 \\
& 128 & 49292 & 698 & -0.115 & 0.005  & 49.6 \\
\midrule
\multirow{3}{*}{OLMo-2-1124-7B-Instruct}
& 32  & 49393 & 88  & 0.880  & -0.215 & 52.3 \\
& 64  & 49393 & 348 & -0.303 & -0.012 & 51.1 \\
& 128 & 49306 & 613 & -0.063 & -0.006 & 50.9 \\
\midrule
\multirow{3}{*}{Falcon-7B-Instruct}
& 32  & 49393 & 50  & -0.463 & -0.199 & 50.0 \\
& 64  & 49393 & 291 & 0.062  & -0.089 & 57.0 \\
& 128 & 49392 & 576 & 0.037  & 0.016  & 48.3 \\
\bottomrule
\end{tabular}
\caption{
Semantic context effect on exposed samples when adding grounding instructions. $S$ denotes the change in suffix log-probability when replacing the comparison context with the relevant retrieved context. The effect is generally inclined towards negative, but less pronounced than the one without the grounding instructions (table \ref{tab:semantic-generic}).
}
\label{tab:semantic-boundary-effect}
\end{table*}

\section{Effect of Chat Templates}
\label{app:chat-template}

Our main experiments focus on the standard prefix--suffix extraction setting, which has been the dominant setup in prior memorization studies. In this setting, the suffix is scored as a direct continuation of the prefix. A natural alternative would be to place the prefix inside a full chat template. However, this introduces a different measurement regime.

As illustrated in Figure~\ref{fig:plain-vs-chat-schematic}, adding a chat template inserts an assistant-role boundary between the prefix and the scored suffix. As a result, for the sequence $(p,s)$ that appeared in training, the model is no longer asked to continue $p$ with $s$, but instead to generate $s$ after an assistant tag. In other words, you would measure the extraction rate of  $(p,\texttt{<assistant>},s)$, which is a different value. This effect has also been noted by \citet{nasr2025scalable}.

Figure~\ref{fig:plain-vs-chat} shows that this effect is large: across all models and prefix lengths, adding the chat template substantially lowers measured memorization. Because this shift is much larger than the contextual effects we study in the main paper, including chat-template formatting would make the results harder to interpret by mixing the effect of preceding context with a different response-formatting regime. For this reason, the main paper isolates the effect of preceding context and does not include chat-template structure in the core analysis.

\begin{figure*}[t]
    \centering
    \small
    \setlength{\fboxsep}{6pt}
    \begin{minipage}{0.47\linewidth}
        \centering
        \textbf{Plain continuation}\\[0.5em]
        \fbox{
        \parbox{0.92\linewidth}{
        \texttt{[prefix]} \\
        $\rightarrow$ score \texttt{[suffix]}
        }}
    \end{minipage}
    \hfill
    \begin{minipage}{0.47\linewidth}
        \centering
        \textbf{Chat template}\\[0.5em]
        \fbox{
        \parbox{0.92\linewidth}{
        \texttt{<|user|> [prefix]} \\
        \texttt{<|assistant|>} \\
        $\rightarrow$ score \texttt{[suffix]}
        }}
    \end{minipage}
    \caption{
    Illustration of plain prefix--suffix scoring versus chat-template scoring. In the plain setting, the suffix is scored as a direct continuation of the prefix. In the chat-template setting, an assistant-role boundary is inserted between the prefix and suffix, changing the continuation being measured.
    }
    \label{fig:plain-vs-chat-schematic}
\end{figure*}

\begin{figure*}[t]
    \centering
    \includegraphics[width=\textwidth]{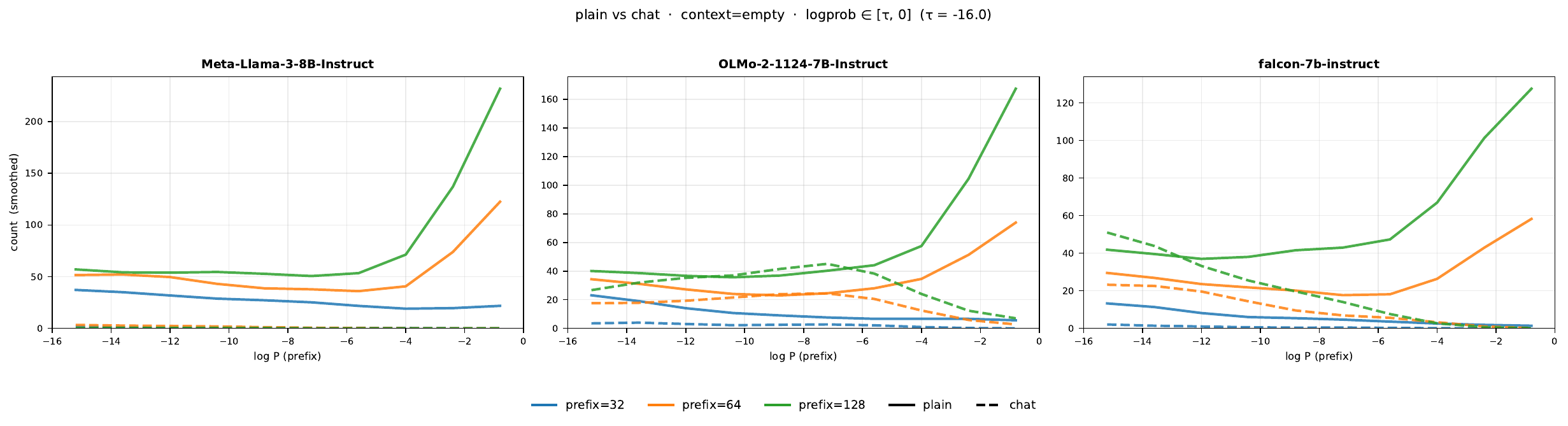}
    \caption{
    Effect of chat-template formatting on measured memorization under empty context. Solid lines show plain prefix--suffix scoring, while dashed lines show chat-template scoring. The large drop under chat-template scoring shows that template formatting has a strong effect on the measurement.
    }
    \label{fig:plain-vs-chat}
\end{figure*}

\section{Detailed Experimental Setup}
\label{app:experimental-setup}

This appendix lists implementation details for reproducibility.

\begin{itemize}
    \item Sampling: top-$p = 1.0$, temperature $T = 1.0$ (scoring
    setting; no actual sampling required since we compute
    log-probabilities directly).
    \item Log-probability thresholding: across all plots the threshold is $-16$.
    \item Suffix length: 32 tokens.
    \item Context window: model maximum.
\end{itemize}





\section{Additional Visualizations}
\label{app:additional-visualizations}

\subsection{Semantic Relevance Effect}
\label{sub:semantic-relevance}
\begin{figure*}[t]
    \centering
    \includegraphics[width=\textwidth]{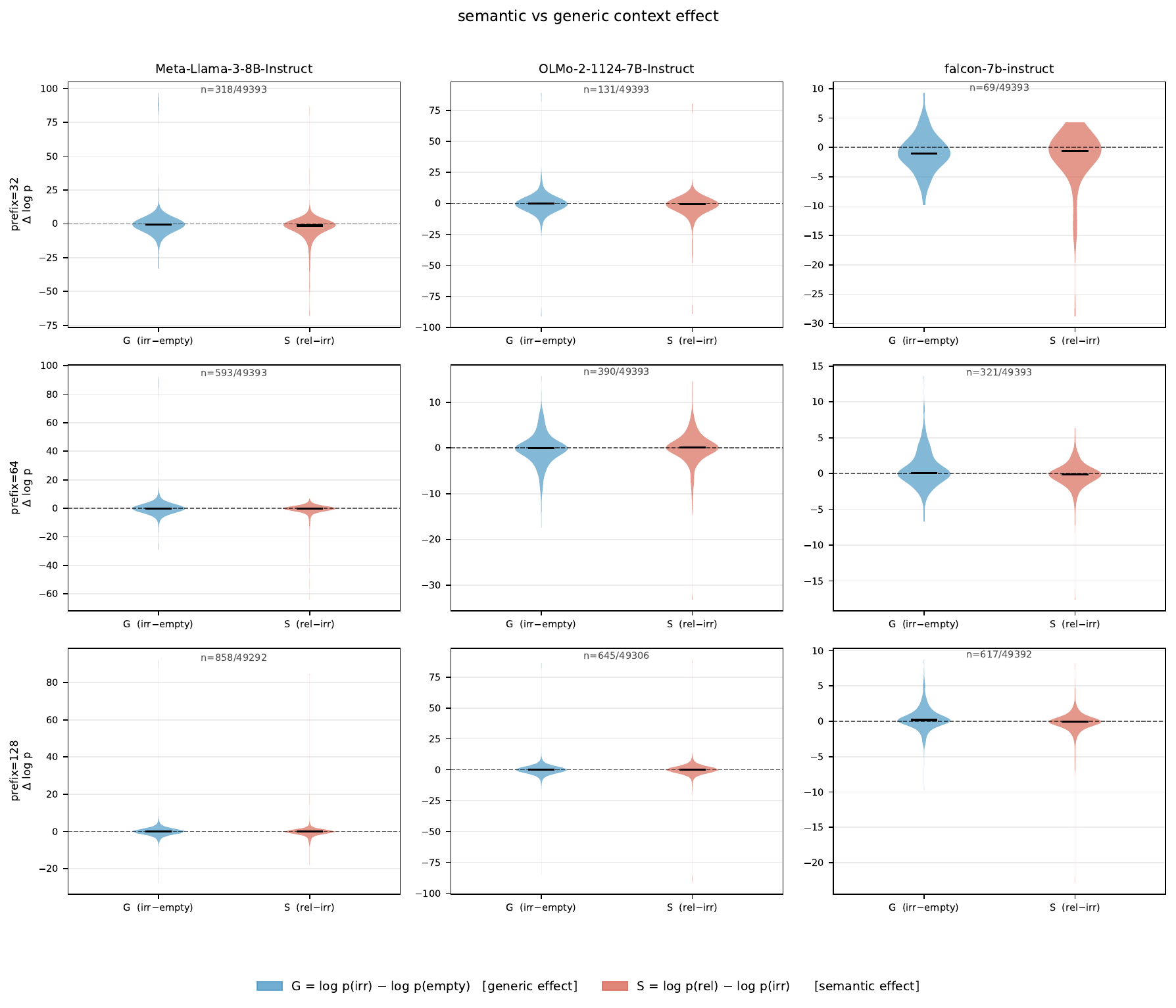}
    \caption{
    Semantic versus generic context effects for exposed samples. Blue distributions show $G=\ell^{\mathrm{irr}}-\ell^{\emptyset}$, the effect of adding irrelevant context. Red distributions show $S=\ell^{\mathrm{rel}}-\ell^{\mathrm{irr}}$, the additional effect of replacing irrelevant context with relevant context. Relevant context is often suppressive, especially for shorter prefixes, but generic contextual perturbation also contributes and is more model-dependent.
    }
    \label{fig:semantic-generic}
\end{figure*}

Figure~\ref{fig:semantic-generic} shows the distributions of $G$ and $S$ for exposed samples (related to section \ref{sec:res-semantic-generic}). The generic effect is model-dependent: irrelevant context can suppress extraction in some settings, while having little effect or increasing extraction in others. In contrast, the semantic relevance effect is often negative, especially for shorter prefixes. This suggests that related context can add a suppressive effect, although this effect weakens as prefix length increases.

\subsection{Relevant versus Irrelevant Context}
\label{app:scatter-rel-irr}

\begin{figure*}[t]
    \centering
    \includegraphics[width=\textwidth]{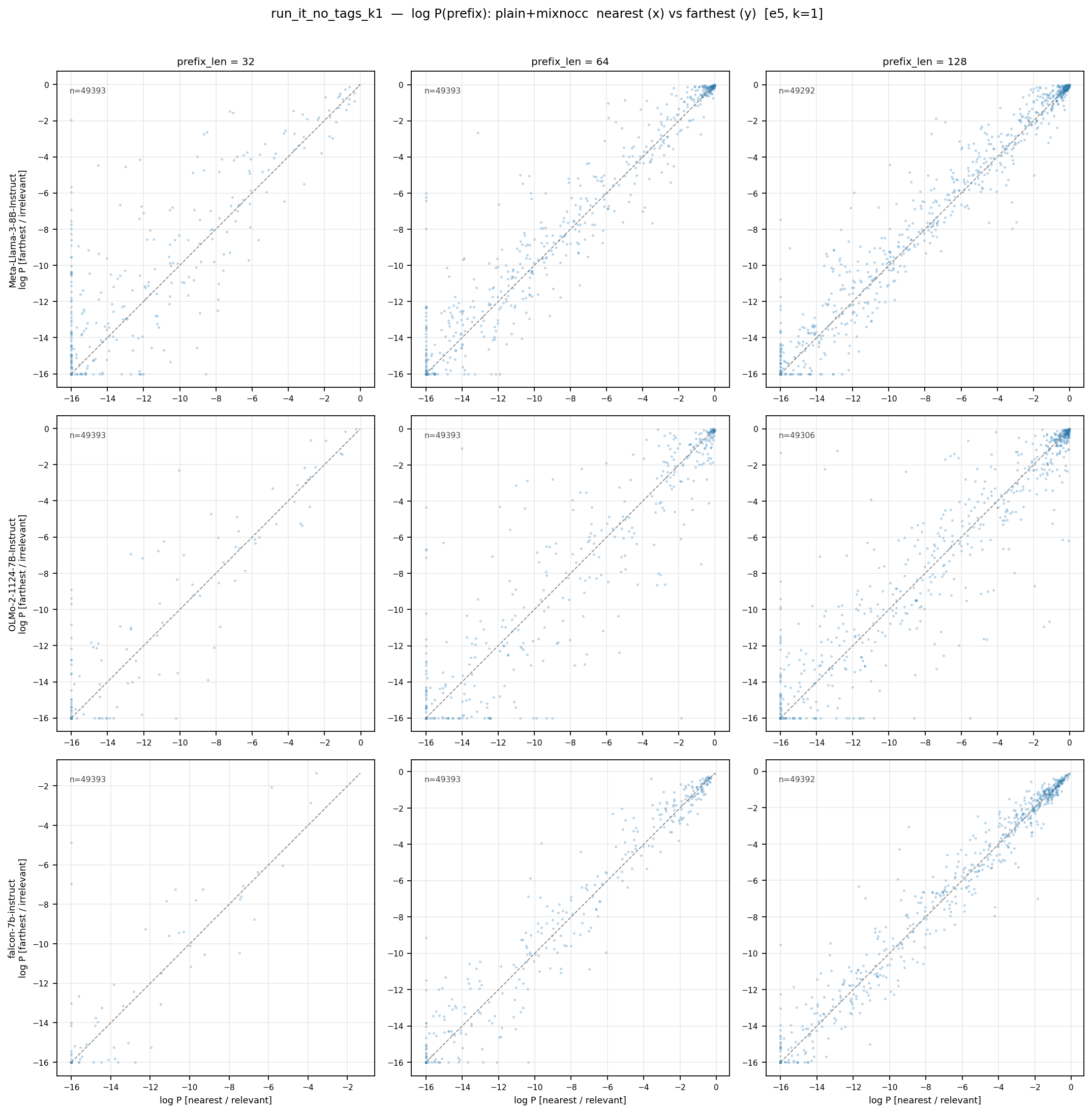}
    \caption{
    Item-level scatter plot of suffix log-probabilities under relevant and irrelevant context. The x-axis shows log-probability under relevant retrieved context, and the y-axis shows log-probability under irrelevant context. Points near the diagonal indicate similar extraction likelihood under both context types; deviations from the diagonal indicate context-sensitive samples.
    }
    \label{fig:scatter-rel-irr}
\end{figure*}

Figure~\ref{fig:scatter-rel-irr} compares suffix log-probabilities under relevant and irrelevant context at the item level. Each point corresponds to one prefix--suffix sample. Points near the diagonal indicate samples whose extraction likelihood is similar under both context types, while deviations from the diagonal indicate samples that are more sensitive to the choice of retrieved context.

%
%
  \sloppy
  \setlength{\emergencystretch}{3em}

\providecommand{\fieldrule}{%
  \arrayrulecolor{black!20}\cmidrule[0.2pt]{1-2}\arrayrulecolor{black}}

\section{Examples of Recalled and Forgotten Samples in Presence of Context}\label{app:examples}
Tables~\ref{tab:forgotten} and~\ref{tab:recalled} show several examples of items
that become un-extractable (forgotten) and extractable (recalled) in the presence
of the retrieved context.

\begin{table*}[t]
\centering
\caption{Forgotten items: sequences that are extracted verbatim by the model without retrieved context, but no longer under context. We show one example per model, restricted to items whose log-probability gap satisfies $|\log p_{\text{ctx}} - \log p_{\text{no-ctx}}| \geq 5$ and where exactly one of the two log-probabilities falls below $-16$, so that the pair straddles the extraction threshold.}
\label{tab:forgotten}
\small
\begin{tabular}{@{}>{\raggedright\arraybackslash}p{0.13\textwidth}>{\raggedright\arraybackslash}p{0.83\textwidth}@{}}
\toprule
\multicolumn{2}{@{}l@{}}{\textbf{Example 1}}\\[2pt]
\textit{Model} & tiiuae/falco\allowbreak{}n-7b-instruc\allowbreak{}t \\
\fieldrule
\textit{Prefix len} & 32 tokens \quad (suffix len: 32) \\
\fieldrule
\textit{Log-probs} & $\log p_{\text{no-ctx}} = -8.680$ \quad $\log p_{\text{ctx}} = -25.069$ \quad $\Delta = -16.389$ \\
\fieldrule
\textit{Prefix} & \small\ttfamily Cambridge, ID Fertilizers \newline \hfill \newline YP - The Real Yellow PagesSM - helps you find the right local businesses to meet your specific needs. Search \\[2pt]
\fieldrule
\textit{Suffix} & \small\ttfamily  results are sorted by a combination of factors to give you a set of choices in response to your search criteria. These factors are similar to those you might use to \\[2pt]
\fieldrule
\textit{Context} & \scriptsize Q: \newline \hfill \newline cms for software project hosting \newline \hfill \newline I want to make a site where I present my own open source software project. It should look like http://www.s\allowbreak{}chillmania.c\allowbreak{}om/projects/\allowbreak{}soundmanager\allowbreak{}2/ for example. \newline Are there any CMS (PHP) with this kind of functionality preconfigured? What I need is code highlighting, code documentation, download statistic, bug tracker, discussion forum, donation button etc. \newline I think if I take a standard CMS like wordpress or drupal I'll spend ages to find the right modules and setup the whole thing. \newline I will also accept answers that recommend a useful drupal distribution or a list of wordpress plugins.. \newline \hfill \newline A: \newline \hfill \newline Have you tried looking at Ohloh or the CMS Matrix? Some of the CMS's descriptions in there haven't been updated in a long time though. (I would have posted it as a comment but I can't just yet and I tought it would still be interesting for you to know.) \\
\bottomrule
\end{tabular}
\end{table*}

\begin{table*}[t]
\centering
\ContinuedFloat
\caption{Forgotten items (continued).}
\small
\begin{tabular}{@{}>{\raggedright\arraybackslash}p{0.13\textwidth}>{\raggedright\arraybackslash}p{0.83\textwidth}@{}}
\toprule
\multicolumn{2}{@{}l@{}}{\textbf{Example 2}}\\[2pt]
\textit{Model} & meta-llama/M\allowbreak{}eta-Llama-3-\allowbreak{}8B-Instruct \\
\fieldrule
\textit{Prefix len} & 32 tokens \quad (suffix len: 32) \\
\fieldrule
\textit{Log-probs} & $\log p_{\text{no-ctx}} = -2.858$ \quad $\log p_{\text{ctx}} = -49.586$ \quad $\Delta = -46.728$ \\
\fieldrule
\textit{Prefix} & \small\ttfamily Tucson craigslist adult - tour cheap \newline \hfill \newline About me: Hello My name is Melly. These factors are similar to those you might use to determine which business \\[2pt]
\fieldrule
\textit{Suffix} & \small\ttfamily  to select from a local Yellow Pages directory, including proximity to where you are searching, expertise in the specific services or products you need, and comprehensive business information to \\[2pt]
\fieldrule
\textit{Context} & \scriptsize Q: \newline \hfill \newline Plotting More than 2 Factors \newline \hfill \newline Suppose I ran a factor analysis \& got 5 relevant factors. Now, I want to graphically represent the loading of these factors on the variables. Can anybody please tell me how to do it. I can do using 2 factors. But can't able to do when number of factors are more than 2. \newline The 2 factor plotting is given in "Modern Applied Statistics with S", Fig 11.13. I want to create similar graph but with more than 2 factors. Please find the snap of the Fig mentioned above: \newline \hfill \newline X \& y axes are the 2 factors. \newline Regards, \newline Ari \newline \hfill \newline A: \newline \hfill \newline Beware: not the answer you are looking for and might be incorrect also, this is my subjective thought. \newline I think you run into the problem of sketching several dimensions on a two dimension screen/paper.  \newline I would say there is no sense in plotting more factors' or PCs' loadings, but if you really insist: display the first two (based on eigenvalues) or create only 2 factors. Or you could reduce dimension by other methods also (e.g. MDS). \newline Displaying 3 factors' loadings in a 3 dimensional graph would be just hardly clear, not to think about more factors. \newline \hfill \newline UPDATE: I had a dream about trying to be more ontopic :) \\
\bottomrule
\end{tabular}
\end{table*}

\begin{table*}[t]
\centering
\ContinuedFloat
\caption{Forgotten items (continued).}
\small
\begin{tabular}{@{}>{\raggedright\arraybackslash}p{0.13\textwidth}>{\raggedright\arraybackslash}p{0.83\textwidth}@{}}
\toprule
\multicolumn{2}{@{}l@{}}{\textbf{Example 3}}\\[2pt]
\textit{Model} & allenai/OLMo\allowbreak{}-2-0425-1B-I\allowbreak{}nstruct \\
\fieldrule
\textit{Prefix len} & 64 tokens \quad (suffix len: 32) \\
\fieldrule
\textit{Log-probs} & $\log p_{\text{no-ctx}} = -6.081$ \quad $\log p_{\text{ctx}} = -17.202$ \quad $\Delta = -11.121$ \\
\fieldrule
\textit{Prefix} & \small\ttfamily Menu \newline Notice: Trying to get property of non-object in /var/www/ipo\allowbreak{}-book.com/pu\allowbreak{}blic\_html/wp\allowbreak{}-content/the\allowbreak{}mes/ipo/func\allowbreak{}tions.php on line 600 \newline Notice: Trying to get property of non-object in /var/www/ipo\allowbreak{}-book.com/pu\allowbreak{}blic\_html/wp\allowbreak{}-content/the\allowbreak{}mes/ipo/func\allowbreak{}tions.php on line 602 \\[2pt]
\fieldrule
\textit{Suffix} & \small\ttfamily Notice: Trying to get property of non-object in /var/www/ipo\allowbreak{}-book.com/pu\allowbreak{}blic\_html/wp\allowbreak{}-content/the\allowbreak{}mes/ipo/func\allowbreak{}tions.php on line 604 \newline Notice \\[2pt]
\fieldrule
\textit{Context} & \scriptsize Q: \newline \hfill \newline ?????? segmentation error \newline \hfill \newline ????? ?????????? ????? ?? ??????? ? ????????? ??????? ?????? segmentation error, ??? ?? ????????? \newline \hfill \newline ??? ??? ????????? ? BookContact.txt \newline Sergey\_Kotom\allowbreak{}kin:154:8(77\allowbreak{}7)777-77-77: \newline Maks\_Loxin:8\allowbreak{}4:8(952)748-\allowbreak{}41-70: \newline Andrey\_Serov\allowbreak{}:14:8(952)74\allowbreak{}7-42-70: \newline Aleks\_Sagin:\allowbreak{}14:8(952)509\allowbreak{}-08-07: \newline Dimon\_Wanow:\allowbreak{}17:8(785)124\allowbreak{}-47-45: \newline \hfill \newline ??? ??? \newline \#include \textless{}stdio.h\textgreater{} //??? printf, fprintf, fopen, fclose \newline \#include \textless{}string.h\textgreater{} \newline \#include \textless{}stdlib.h\textgreater{} \newline \hfill \newline \#define FNAME "BookContact.txt" //????????? ???? ??? ???????? \newline \hfill \newline void startMenu()\{ //? ??? ???????? ????????? ???? \newline \hfill \newline     printf("----\allowbreak{}------------\allowbreak{}-----Main Menu--------\allowbreak{}------------\allowbreak{}-\textbackslash{}n"); \newline     printf("This is the phone book of the company 'YptRemont'\textbackslash{}n"); \\
\bottomrule
\end{tabular}
\end{table*}

\begin{table*}[t]
\centering
\ContinuedFloat
\caption{Forgotten items (continued).}
\small
\begin{tabular}{@{}>{\raggedright\arraybackslash}p{0.13\textwidth}>{\raggedright\arraybackslash}p{0.83\textwidth}@{}}
\toprule
\multicolumn{2}{@{}l@{}}{\textbf{Example 4}}\\[2pt]
\textit{Model} & allenai/OLMo\allowbreak{}-2-1124-7B-I\allowbreak{}nstruct \\
\fieldrule
\textit{Prefix len} & 32 tokens \quad (suffix len: 32) \\
\fieldrule
\textit{Log-probs} & $\log p_{\text{no-ctx}} = -7.669$ \quad $\log p_{\text{ctx}} = -20.781$ \quad $\Delta = -13.112$ \\
\fieldrule
\textit{Prefix} & \small\ttfamily Personal Income and Outlays: September 2009 \newline \hfill \newline Personal income decreased \$0.1 billion, or less than 0.1 percent, and disposable personal \\[2pt]
\fieldrule
\textit{Suffix} & \small\ttfamily  income (DPI) decreased \$0.2 billion, or less than 0.1 percent, in September, according to the Bureau of Economic Analysis. \\[2pt]
\fieldrule
\textit{Context} & \scriptsize Since the completion of the Trans @-@ Alaska Pipeline System in 1977 , the government of the state of Alaska has been reliant on taxes paid by oil producers and shippers . Prior to 1976 , Alaska 's personal income tax rate was 14 @.@ 5 percent --- the highest in the United States . The gross state product was \$ 8 billion , and Alaskans earned \$ 5 billion in personal income . Thirty years after the pipeline began operating , the state had no personal income tax , the gross state product was \$ 39 billion , and Alaskans earned \$ 25 billion in personal income . Alaska moved from the most heavily taxed state to the most tax @-@ free state . \\
\bottomrule
\end{tabular}
\end{table*}
\clearpage

\clearpage

\begin{table*}[t]
\centering
\caption{Recalled items: sequences that are not extracted by the model alone, but become extractable once retrieved context is prepended. We show one example per model, restricted to items whose log-probability gap satisfies $|\log p_{\text{ctx}} - \log p_{\text{no-ctx}}| \geq 5$ and where exactly one of the two log-probabilities falls below $-16$, so that the pair straddles the extraction threshold.}
\label{tab:recalled}
\small
\begin{tabular}{@{}>{\raggedright\arraybackslash}p{0.13\textwidth}>{\raggedright\arraybackslash}p{0.83\textwidth}@{}}
\toprule
\multicolumn{2}{@{}l@{}}{\textbf{Example 1}}\\[2pt]
\textit{Model} & tiiuae/falco\allowbreak{}n-7b-instruc\allowbreak{}t \\
\fieldrule
\textit{Prefix len} & 64 tokens \quad (suffix len: 32) \\
\fieldrule
\textit{Log-probs} & $\log p_{\text{no-ctx}} = -20.885$ \quad $\log p_{\text{ctx}} = -5.609$ \quad $\Delta = +15.277$ \\
\fieldrule
\textit{Prefix} & \small\ttfamily Game seems to be working just fine, but trainer is not. I see that there might have been some patches? How do i check my game version? Trainer wont work without the right version haha so need to check wich version i need to get my game too. \newline \hfill \newline Game seems to be working just fine, \\[2pt]
\fieldrule
\textit{Suffix} & \small\ttfamily  but trainer is not. I see that there might have been some patches? How do i check my game version? Trainer wont work without the right version haha so \\[2pt]
\fieldrule
\textit{Context} & \scriptsize Special abilities and actions are controlled by the Force Gauge , an energy meter which determines how long special skills can be active . Once drained , the skill cannot be activated until the meter refills , which it does during battles . Four skills can be equipped for each character . After battles , characters earn Enhancement Points ( EP ) , which can be used to strengthen equipped armor , weapons and accessories , instead of using Gald ( the in @-@ game currency ) as with previous Tales titles . EP can be transferred between weapons . An optional battle arena was added in the PSP version of the game with both single @-@ player and multiplayer matches : in both modes , players need to defeat an enemy team ( either AI @-@ controlled enemies or an opposing team ) within a time limit . \\
\bottomrule
\end{tabular}
\end{table*}

\begin{table*}[t]
\centering
\ContinuedFloat
\caption{Recalled items (continued).}
\small
\begin{tabular}{@{}>{\raggedright\arraybackslash}p{0.13\textwidth}>{\raggedright\arraybackslash}p{0.83\textwidth}@{}}
\toprule
\multicolumn{2}{@{}l@{}}{\textbf{Example 2}}\\[2pt]
\textit{Model} & meta-llama/M\allowbreak{}eta-Llama-3-\allowbreak{}8B-Instruct \\
\fieldrule
\textit{Prefix len} & 32 tokens \quad (suffix len: 32) \\
\fieldrule
\textit{Log-probs} & $\log p_{\text{no-ctx}} = -21.231$ \quad $\log p_{\text{ctx}} = -8.811$ \quad $\Delta = +12.420$ \\
\fieldrule
\textit{Prefix} & \small\ttfamily Thou still unravish' d bride of quietness, \newline Thou foster-child of silence and slow time, \newline Sylvan historian, who canst thus \\[2pt]
\fieldrule
\textit{Suffix} & \small\ttfamily  express \newline A flowery tale more sweetly than our rhyme: \newline What leaf-fring'd legend haunts about thy shape \newline Of deities or mortals, \\[2pt]
\fieldrule
\textit{Context} & \scriptsize Q: \newline \hfill \newline Word construction like "philanthropist" but regarding silence instead of humankind? \newline \hfill \newline What would be the Latin/English term for somebody who loves silence? Featuring phil(e) as pre- or suffix. \newline Like ... silenciophile? That maybe makes "some sense" - but I've no idea if it is correct or if there would be a canonical neologism. \newline \hfill \newline A: \newline \hfill \newline Latin freely adopted Greek words, including compounds, but in general did not freely create compounds with both Latin and Greek elements, like the modern creation "sociology", and similarly with your suggested form "silenciophile" (better "silentiophile").  If you want to create a compound with "-phile", then "hesychophile" would be better, consisting of Greek elements only. \newline \hfill \newline --- \newline \hfill \newline You made me very low @-@ spirited last night , by your manner of talking --- You are my only friend --- the only person I am intimate with . --- I never had a father , or a brother --- you have been both to me , ever since I knew you --- yet I have sometimes been very petulant . --- I have been thinking of those instances of ill @-@ humour and quickness , and they appeared like crimes . Yours sincerely , Mary . \newline \hfill \newline --- \newline \hfill \newline being desirous of rendering it easy for the future , have made it my stud [...] \\
\bottomrule
\end{tabular}
\end{table*}

\begin{table*}[t]
\centering
\ContinuedFloat
\caption{Recalled items (continued).}
\small
\begin{tabular}{@{}>{\raggedright\arraybackslash}p{0.13\textwidth}>{\raggedright\arraybackslash}p{0.83\textwidth}@{}}
\toprule
\multicolumn{2}{@{}l@{}}{\textbf{Example 3}}\\[2pt]
\textit{Model} & allenai/OLMo\allowbreak{}-2-0425-1B-I\allowbreak{}nstruct \\
\fieldrule
\textit{Prefix len} & 128 tokens \quad (suffix len: 32) \\
\fieldrule
\textit{Log-probs} & $\log p_{\text{no-ctx}} = -16.620$ \quad $\log p_{\text{ctx}} = -8.394$ \quad $\Delta = +8.226$ \\
\fieldrule
\textit{Prefix} & \small\ttfamily Model Method - Questions and Answers \newline \hfill \newline Question posted by Yvonne from Sri Lanka: \newline \hfill \newline Grade/Level: 4th \newline \hfill \newline Question solved by Model Method: 900 people attended a party. 1/4 of the males and 1/7 of the females were children. the number of male and female adults were the same. how many women were there at the party? \newline \hfill \newline Answer: \newline \hfill \newline Step 1: This question involves the Equal Concept. Since 1/4 of the males were children, then 3/4 of the males must be adults. We draw 3 boxes to represent the total number of male adults and 1 \\[2pt]
\fieldrule
\textit{Suffix} & \small\ttfamily  box to represent the total number of male children. \newline \hfill \newline Step 2: Since 1/7 of the females were children, then 6/7 of the \\[2pt]
\fieldrule
\textit{Context} & \scriptsize Q: \newline \hfill \newline Inclusion-Exclusion Principle for basic combinatorics problem... \newline \hfill \newline How many ways are there to pick five people for a committee if there are six \newline   (different) men and eight (different) women and the selection must include at \newline   least one man and one woman? \newline \hfill \newline I know to solve this with very basic combinatorial analysis (though I know the principle is also basic). We find the total people we can choose for the committee and proceed with cases: 1 where we have all men on the committee, 1 where we have all women on the committee and then we subtract the two cases from our total. This gives us \newline \$\$\{14 \textbackslash{}choose 5\} - \{8 \textbackslash{}choose 5\}-\{6 \textbackslash{}choose 5\}.\$\$ \newline How would I compute for Inclusion-Exclusion Principle, though? I've no idea where to begin with it other drawing a venn diagram and labeling set M and W... \newline \hfill \newline A: \newline \hfill \newline You have more-or-less done it already. \newline For \$i \textbackslash{}in \textbackslash{}\{0,1,\textbackslash{}ldots\textbackslash{}\}\$, let \$a\_i\$ be the number of committees with \$\textbackslash{}geq i\$ absent genders. \newline We include \$a\_0=\textbackslash{}binom\{14\}\{5\}\$. \newline We exclude \$a\_1=\textbackslash{}binom\{\allowbreak{}8\}\{5\}+\textbackslash{}binom\allowbreak{}\{6\}\{5\}\$. \newline We include \$a\_2=0\$. \newline We exclude \$a\_3=0\$. \\
\bottomrule
\end{tabular}
\end{table*}

\begin{table*}[t]
\centering
\ContinuedFloat
\caption{Recalled items (continued).}
\small
\begin{tabular}{@{}>{\raggedright\arraybackslash}p{0.13\textwidth}>{\raggedright\arraybackslash}p{0.83\textwidth}@{}}
\toprule
\multicolumn{2}{@{}l@{}}{\textbf{Example 4}}\\[2pt]
\textit{Model} & allenai/OLMo\allowbreak{}-2-1124-7B-I\allowbreak{}nstruct \\
\fieldrule
\textit{Prefix len} & 64 tokens \quad (suffix len: 32) \\
\fieldrule
\textit{Log-probs} & $\log p_{\text{no-ctx}} = -28.025$ \quad $\log p_{\text{ctx}} = -3.848$ \quad $\Delta = +24.177$ \\
\fieldrule
\textit{Prefix} & \small\ttfamily \%PDF-1.2 \newline \%âãÏÓ \newline 1 0 obj\textless{}\textgreater{}endobj \newline 2 0 obj\textless{}\textgreater{}endobj \newline 3 0 obj\textless{}\textgreater{}endobj \newline 4 0 obj\textless{}\textgreater{}endobj \newline 5 0 obj\textless{}\textgreater{}endobj \newline 6 0 obj\textless{}\textgreater{}endobj \newline 7  \\[2pt]
\fieldrule
\textit{Suffix} & \small\ttfamily 0 obj\textless{}\textgreater{}endobj \newline 8 0 obj\textless{}\textgreater{}endobj \newline 9 0 obj\textless{}\textgreater{}endobj \newline 10 0 obj\textless{}\textgreater{}endobj \newline 11  \\[2pt]
\fieldrule
\textit{Context} & \scriptsize Q: \newline \hfill \newline Does anybody use Pancake? MathJax \newline \hfill \newline \$\$\textbackslash{}begin\{ali\allowbreak{}gn*\}a\_1=\textbackslash{}alp\allowbreak{}ha \_1+\textbackslash{}alpha \_2+\textbackslash{}cdots+\textbackslash{}alpha \_n,a\_2=\textbackslash{}alpha \_1\textbackslash{}alpha \_2+\textbackslash{}alpha \_1\textbackslash{}alpha \_3+\textbackslash{}cdots+\textbackslash{}alpha \_\{n-1\}\textbackslash{}alpha \_n,\textbackslash{}cdots a\_1 \textbackslash{}end\{align*\}\$\$ \newline \hfill \newline Delete a\_1 in the above, then that could be shown well. \newline \$\$\textbackslash{}begin\{ali\allowbreak{}gn*\}a\_1=\textbackslash{}alp\allowbreak{}ha \_1+\textbackslash{}alpha \_2+\textbackslash{}cdots+\textbackslash{}alpha \_n,a\_2=\textbackslash{}alpha \_1\textbackslash{}alpha \_2+\textbackslash{}alpha \_1\textbackslash{}alpha \_3+\textbackslash{}cdots+\textbackslash{}alpha \_\{n-1\}\textbackslash{}alpha \_n,\textbackslash{}cdots \textbackslash{}end\{align*\}\$\$ \newline \hfill \newline http://hyper\allowbreak{}groups.panca\allowbreak{}keapps.com/T\allowbreak{}est/test.txt \newline Of course they can show well in Math.SE \newline \$\$\textbackslash{}begin\{ali\allowbreak{}gn*\}a\_1=\textbackslash{}alp\allowbreak{}ha \_1+\textbackslash{}alpha \_2+\textbackslash{}cdots+\textbackslash{}alpha \_n,a\_2=\textbackslash{}alpha \_1\textbackslash{}alpha \_2+\textbackslash{}alpha \_1\textbackslash{}alpha \_3+\textbackslash{}cdots+\textbackslash{}alpha \_\{n-1\}\textbackslash{}alpha \_n,\textbackslash{}cdots a\_1 \textbackslash{}end\{align*\}\$\$ \newline \hfill \newline --- \newline \hfill \newline Q: \newline \hfill \newline ??????? ????? "???????????" ???????? \newline \hfill \newline ???? ???????? ????, ?? ??????? ??? ???? 2 ???????, ??????? ???? ??????? ? ?????????????. ???????? ???????? ?? 2 ?????????? "div" ? "div2" - ?????????????? ??? ?? ? id ???? ???? ????????. \newline \hfill \newline --- \newline \hfill \newline Q: \newline \hfill \newline Order the growth rate from slowest to fastest \newline \hfill \newline I am relatively new to Big-O notation and I came across this question:   \newline \hfill \newline Sort the following functions by order of growth from slowest to fastest - Big-O Notation. For each pair of adjacent functions in your list, please write a sentence describing why it is or [...] \\
\bottomrule
\end{tabular}
\end{table*}
\clearpage

\clearpage

\end{document}